\PassOptionsToPackage{usenames,dvipsnames,table}{xcolor}
\documentclass[journal,transmag]{IEEEtran}

\usepackage{fixltx2e}
\usepackage{yfonts}
\usepackage{dblfloatfix}
\usepackage{cite}
\usepackage[cmex10]{amsmath}
\usepackage{times}
\usepackage{epsfig}
\usepackage{graphicx}
\usepackage{amsmath}
\usepackage{amssymb}

\usepackage{pgfplotstable}

\usepackage{url}
\usepackage{pbox}

\usepackage{caption}
\usepackage{subcaption}
\usepackage{algorithm,algpseudocode}
\usepackage{rotating}
\usepackage{multirow}
\usepackage{arydshln}
\usepackage{makecell}

\usepackage{booktabs}
\usepackage{pifont}
\usepackage{siunitx}
\usepackage{booktabs,colortbl,array}
\usepackage[inline]{enumitem}
\usepackage{tikz}
\usepackage{adjustbox}
\usepackage[11pt]{moresize}
\usetikzlibrary{shapes.arrows, fadings}

\usepackage{tablefootnote}

\usepackage[pagebackref=true,breaklinks=true,letterpaper=true,colorlinks,bookmarks=false]{hyperref}

\usepackage{pdfpages}

\def\overfeat{\texttt{OverFeat}}
\def\alexnet{\texttt{AlexNet}}

\def\vgg{\texttt{VGGNet}}
\def\googlenet{\texttt{GoogleNet}}
\def\overfeatA{\texttt{OverFeat} }
\def\alexnetA{\texttt{AlexNet} }

\def\caffeA{\texttt{Caffe} }
\def\L2{$L2$}

\newcommand{\tickYes}{\checkmark}
\newcommand{\tickNo}{\hspace{1pt}\ding{55}}

\definecolor{rulecolor}{RGB}{0,71,171}
\definecolor{tableheadcolor}{gray}{0.92}
\definecolor{darkgreen}{rgb}{0, 0.4, 0}
\definecolor{darkcyan}{rgb}{0, 0.7, 0.7}
\definecolor{lightred}{rgb}{0.9, 0, 0}
\definecolor{darkpink}{rgb}{1, 0, 0.5}

\newcommand{\topline}{ %
        \arrayrulecolor{rulecolor}\specialrule{0.1em}{\abovetopsep}{0pt}%
        \arrayrulecolor{tableheadcolor}\specialrule{\belowrulesep}{0pt}{0pt}%
        \arrayrulecolor{rulecolor}}
\newcommand{\midtopline}{ %
        \arrayrulecolor{tableheadcolor}\specialrule{\aboverulesep}{0pt}{0pt}%
        \arrayrulecolor{rulecolor}\specialrule{\lightrulewidth}{0pt}{0pt}%
        \arrayrulecolor{white}\specialrule{\belowrulesep}{0pt}{0pt}%
        \arrayrulecolor{rulecolor}}
\newcommand{\bottomline}{ %
        \arrayrulecolor{white}\specialrule{\aboverulesep}{0pt}{0pt}%
        \arrayrulecolor{rulecolor} %
        \specialrule{\heavyrulewidth}{0pt}{\belowbottomsep}}%

\newcommand{\captionfontsize}[0]{%
        \small}

\tikzfading[name=fade right,right color=transparent!0, left
  color=transparent!70]
\tikzfading[name=fade left, left color=transparent!0, right color=transparent!70]

\def\egA{{\emph{e.g.}} }

\newif\ifreview

\reviewfalse

\begin{document}
\title{Factors of Transferability for a Generic ConvNet Representation}


\author{\IEEEauthorblockN{Hossein Azizpour, 
Ali Sharif Razavian, 
Josephine Sullivan, 
Atsuto Maki,
Stefan Carlsson\vspace{0.3cm}}{\small \{azizpour, razavian, sullivan, atsuto, stefanc\}@csc.kth.se}
\IEEEauthorblockA{Computer Vision and Active Perception (CVAP),
Royal Institute of Technology (KTH), Stockholm, SE-10044 Sweden \vspace{0.24cm}}

\thanks{The manuscript is an extended version of the CVPR Workshops DeepVision 2015 paper "From Generic to Specific Deep Representations for Visual Recognition" . \ifreview The CVPR paper is attached at the end of the manuscript. \else \fi}}

%

\IEEEtitleabstractindextext{%
\begin{abstract}
\noindent \begin{abstract}
Evidence is mounting that Convolutional Networks (ConvNets) are the most effective representation
learning method for visual recognition tasks. In the common scenario, a ConvNet is
trained on a large labeled dataset (source) and the feed-forward units
activation of the trained network, at a certain layer of the network, is used as a generic
representation of an input image for a task with relatively smaller training set (target). Recent studies have shown this form
of representation transfer to be suitable for a wide range of
target visual recognition tasks. This paper introduces and investigates several factors affecting the transferability of such
representations. It includes parameters for
training of the source ConvNet such as its architecture, distribution of the training data, etc. and also the parameters of
feature extraction such as layer of the trained ConvNet, dimensionality reduction, etc.  Then, by
optimizing these factors, we show that significant improvements can be achieved on various (17) visual recognition tasks. We further show that these visual recognition tasks can be categorically ordered based on their distance from the
source task such that a correlation between the performance of tasks and their distance from
the source task w.r.t. the proposed factors is observed.
\end{abstract}

\end{abstract}

}

\maketitle

\IEEEdisplaynontitleabstractindextext

\IEEEpeerreviewmaketitle

\section{Introduction}
\label{sec:intro}

\PARstart{C}{onvolutional networks} (ConvNets) trace back to the early
works on digit and character recognition
\cite{Fukushima80,LeCun98}. Prior to 2012, though, in computer vision
field, neural networks were more renowned for their propensity to
overfit than for solving difficult visual recognition problems. And
within the computer vision community it would have been considered
unreasonable, given the overfitting problem, to think that they could be
used to train image representations for transfer learning.

However, these perceptions have had to be radically altered by the
experimental findings of the last three years. First, deep networks
\cite{Krizhevsky12,Girshick13}, trained using large labelled datasets
such as ImageNet \cite{ilsvrc-2013}, produce by a huge margin the best
results on the most challenging image classification
\cite{ilsvrc-2013} and detection datasets \cite{voc-2012}. Second,
these deep ConvNets learn powerful generic image representations
\cite{Razavian14,Donahue14,Oquab13,Zeiler13} which can be used
off-the-shelf to solve many visual recognition problems
\cite{Razavian14}. In fact the performance of these representations is
so good that at this juncture in computer vision, a \textit{deep
  ConvNet image representation} combined with a \textit{simple
  classifier} \cite{Razavian14,Girshick13} should be the first
alternative to try for solving a visual recognition task.
\begin{figure}[t!]
   \centering      
   \includegraphics[width=1\linewidth]{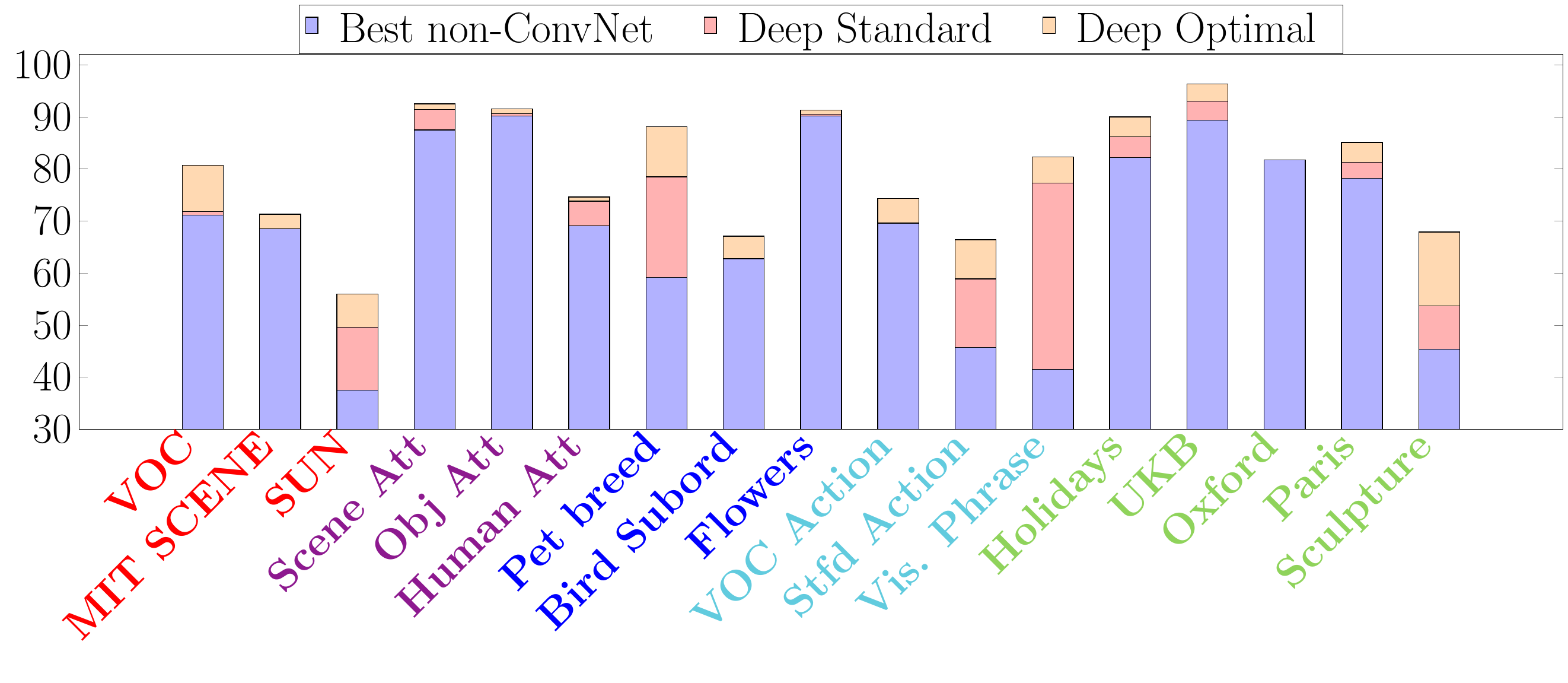}
\caption{\captionfontsize \textbf{The improvements achieved by optimizing
    the transferability factors are significant.}  This optimization
  boosted the performance by up to \textbf{50\%} relative error
  reduction. The violet bars show the performance of non-ConvNet state
  of the art systems on different datasets. The pink stacked bars
  shows the improvement when using off-the-shelf ConvNet features with
  standard settings and a linear SVM classifier. The burnt orange
  stacked bars show the boost gained by finding the best
  transferability factors for each task. Detailed results can be found
  in Table \ref{tab:final_results}. The accuracy is measured using the
  standard evaluation criteria of each task, see the references in
  Table \ref{tab:tasks}.}
\label{fig:teaser_bars}
\end{figure}


  
An \textit{elaborate} classification model based on a 
generic ConvNet representation has sometimes been shown to improve the performance of a simple classifier \cite{Toshev14,Zhang14ECCV} and some other times not so significantly \cite{KarpathyTSLSF14, Girshick15}. In any case, the field has observed that a better ConvNet representation (\egA \vgg~\cite{Simonyan14} or \googlenet~\cite{Szegedy14} instead of \alexnet~\cite{Krizhevsky12}) usually gives more boost in the final performance than a more elaborately designed classification model \cite{Krause15, Girshick13, NgHVVMT15}.

 \begin{figure*}[t]
\centering
  \begin{tabular}{@{}c@{}}
\centering
    \includegraphics[width=\linewidth]{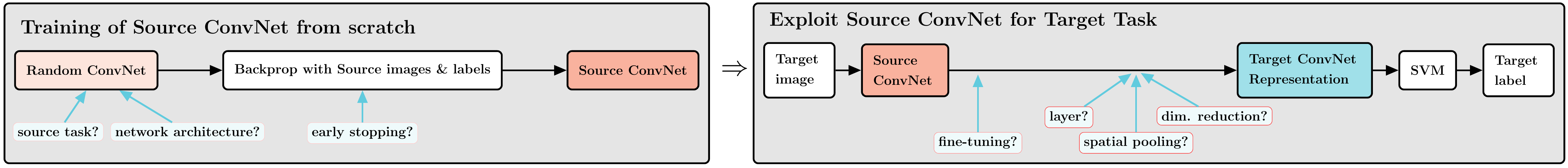}
  \end{tabular}
  \vspace{-0.4cm}
  \caption[]{{\captionfontsize \textbf{Transferring a ConvNet
      Representation} ConvNet representations are effective for visual
    recognition. The picture above shows the pipeline of transferring
    a source ConvNet representation to a target task of interest. We
    define several factors which control the transferability of such
    representations to different tasks (questions with blue
    arrow). These factors come into play at different stages of
    transfer. Optimizing these factors is crucial if one wants to
    maximize the performance of the transferred representation (see
    Figure \ref{fig:teaser_bars}).}}
  \label{fig:problem}
\end{figure*}

Following these observations, a relevant question is: \textit{How can I then maximize the
   performance of the ConvNet representation for my particular
   target task?}  The question becomes especially pertinent if you
 only have a limited amount of labelled training data, time and
 computational resources because training a specialized deep ConvNet
 from scratch is not an option.  The question rephrased in more
 technical terminology is: how should a deep ConvNet representation be
 learned and adjusted to allow better transfer learning from a source
 task producing a generic representation to a specific target task? In
 this paper we identify the relevant factors and demonstrate, from
 experimental evidence, how they should be set given the
 categorization of the target task.\\\\
 
The first set of factors that effect the transferability of a ConvNet
representation are those defining the architecture and training of the
initial deep ConvNet.  These include the source task (encoded in the
labelled training data), network width and depth, distribution of the training data, optimization
parameters. The next set, after learning the ``raw"
representation, are what we term post-learning parameters.  These
include whether you fine-tune the network using labelled data
from the target task, the network layer from which the representation is extracted
and whether the representation should be post-processed by
spatial pooling and dimensionality reduction. 

Figure \ref{fig:problem} gives a graphical overview of how we
transfer a ConvNet representation trained for a source task to target
task and the factors we consider that affect its transferability and
at what stage in the process the factors are applied. While Figure
\ref{fig:teaser_bars} shows how big a difference an optimal
configuration for these factors can make for 17 different target
tasks. 

How should you set these factors? Excitingly we observe that often there is
a pattern for these factors. Their optimal settings are correlated with the distance of the target task's distance
from the source task.  When occasionally there is an exception to the
general pattern there is a plausible explanation.  Table
\ref{tab:best_practice} lists some of our
findings
, driven by our quantitative results, and shows the best
settings for the factors we consider and illustrates the correlations
we mention. 

To summarize deep ConvNet representation are very amenable to transfer
learning. The concrete evidence we present for this assessment is that
in 16 out of 17 diverse standard computer vision databases the
approach just described, based on a deep ConvNet representation
trained with ImageNet and optimal settings of the transferability
settings, outperforms all published non-ConvNet based methods, see
Table \ref{tab:final_results}.

\renewcommand{\arraystretch}{1.0}
\begin{table}
\begin{adjustbox}{max width=\textwidth}
\begin{tabular}{@{}>{\footnotesize}b{.21\linewidth}*{5}{@{}>{\footnotesize}b{.16\linewidth}@{}}}
  \toprule
  & \multicolumn{5}{m{.75\linewidth}}{\centering \textbf{\small Target task}}\\
  \cmidrule{2-6}
  \textbf{\small Factor} & 
   {Source task ImageNet} 
  &\multicolumn{1}{m{.12\linewidth}}{\centering $\cdots$}
  & { FineGrained recognition}
  &\multicolumn{1}{m{.12\linewidth}}{\centering $\cdots$}
  &  {Instance \textcolor{white}{xx} retrieval}\\
  \midrule
 {Early stopping} &
 \multicolumn{5}{m{.75\linewidth}}{
   \adjustbox{valign=m}{\begin{tikzpicture}[trim left=0cm,trim
         right=\linewidth]
     \draw[white, very thin] (0,-8pt) rectangle (\linewidth,8pt);
     \draw[ultra thick,red] (0,0) -- (\linewidth,0) node[inner ysep=0pt]
          [midway,fill=white,font=\footnotesize,draw=white,thin,text=blue!80,anchor=mid] {Don't do it};
   \end{tikzpicture}
}
}
\\[5pt]
  {Network depth} &
 \multicolumn{5}{m{.75\linewidth}}{
   \adjustbox{valign=m}{\begin{tikzpicture}[trim left=0cm,trim
         right=\linewidth]
     \draw[white, very thin] (0,-8pt) rectangle (\linewidth,8pt);
     \draw[ultra thick,red] (0,0) -- (\linewidth,0) node
          [midway,fill=white,font=\footnotesize,draw=white,thin,text=blue!80,anchor=mid,inner
            ysep=0pt] {As deep as possible};
   \end{tikzpicture}
}
}
\\[5pt]
  {Network width} &
  \multicolumn{5}{m{.75\linewidth}}{
   \adjustbox{valign=m}{\begin{tikzpicture}[trim left=0cm,trim right=\linewidth]
     \draw[red,ultra thick,-,path fading=fade left] (0,0) -- (\linewidth-1pt,0)
     ++(0,1pt) ++(0,-1pt);
     \draw[white, very thin] (0,-8pt) rectangle (\linewidth,8pt);
     \draw[red!30,ultra thick,->] (\linewidth-1pt,0) --
     (\linewidth,0);
     \draw[ultra thick, -,draw opacity=0] (0,0) -- (\linewidth,0) node
          [pos=.2,fill=white,font=\footnotesize,draw=white,thin,text=blue!80,anchor=mid,
          inner ysep=0pt]
          {Wider}
          node [pos=.75,fill=white,font=\footnotesize,draw=white,thin,text=blue!80,anchor=mid,inner ysep=0pt]
          {Moderately wide};
   \end{tikzpicture}
  }
}\\[5pt]
  {Diversity/Density} &
 \multicolumn{5}{m{.75\linewidth}}{
   \adjustbox{valign=m}{\begin{tikzpicture}[trim left=0cm,trim
         right=\linewidth]
     \draw[white, very thin] (0,-8pt) rectangle (\linewidth,8pt);
     \draw[ultra thick,red] (0,0) -- (\linewidth,0) node
          [midway,fill=white,font=\footnotesize,draw=white,thin,text=blue!80,anchor=mid,inner
            ysep=0pt] {More classes better than more images per class};
   \end{tikzpicture}
}
}
  \\[5pt]
  {Fine-tuning}&
  \multicolumn{5}{m{.75\linewidth}}{
    \adjustbox{valign=m}{\begin{tikzpicture}[trim left=0cm,trim
          right=\linewidth]
     \draw[white, very thin] (0,-8pt) rectangle (\linewidth,8pt);
      \draw[ultra thick,red] (0,0) -- (\linewidth,0) node[inner ysep=0pt]
           [midway,fill=white,font=\footnotesize,draw=white,thin,text=blue!80,anchor=mid] {Yes,
             more improvement with more labelled data};
    \end{tikzpicture}
}
  }
  \\[5pt]
  {Dim. reduction} &
 \multicolumn{5}{m{.75\linewidth}}{
   \adjustbox{valign=m}{\begin{tikzpicture}[trim left=0cm,trim right=\linewidth]
     \draw[red,ultra thick,-,path fading=fade left] (0,0) -- (\linewidth-1pt,0)
     ++(0,1pt) ++(0,-1pt);
     \draw[white, very thin] (0,-8pt) rectangle (\linewidth,8pt);
     \draw[red!30,ultra thick,->] (\linewidth-1pt,0) -- (\linewidth,0);
     \draw[ultra thick, -,draw opacity=0] (0,0) -- (\linewidth,0) node
          [pos=0.15,fill=white,font=\footnotesize,draw=white,thin,text=blue!80,anchor=mid,inner ysep=0pt]
          {Original dim}
          node
          [pos=0.80,fill=white,font=\footnotesize,draw=white,thin,text=blue!80,anchor=mid,inner ysep=0pt]
          {Reduced dim};
   \end{tikzpicture}
}
}
\\[5pt]
  {Rep. layer}  &
  \multicolumn{5}{m{.75\linewidth}}{
   \adjustbox{valign=m}{\begin{tikzpicture}[trim left=0cm,trim right=\linewidth]
     \draw[red,ultra thick,-,path fading=fade right] (0,0) -- (\linewidth-1pt,0)
     ++(0,1pt) ++(0,-1pt);
     \draw[white, very thin] (0,-8pt) rectangle (\linewidth,8pt);
     \draw[red,ultra thick,->] (\linewidth-1pt,0) -- (\linewidth,0);
     \draw[thin, -,draw opacity=0] (0,0) -- (\linewidth,0) node
          [pos=.15,fill=white,font=\footnotesize,draw=white,thin,text=blue!80,anchor=mid,inner ysep=0pt]
          {Later layers}
          node[pos=.8,fill=white,font=\footnotesize,draw=white,thin,text=blue!80,anchor=mid,
         inner ysep=0pt]
          {Earlier layers};
   \end{tikzpicture}
}
  }
  \\
  \bottomrule
\end{tabular}
\end{adjustbox}
\caption{\captionfontsize \textbf{Best practices} to transfer a ConvNet
  representation trained for the source task of ImageNet to a target tasks summarizing some of our findings. The target tasks above are listed from left to right
  according to their increased distance from the source task (ImageNet
  image classification). The table summarizes qualitatively the best
  setting for each factor affecting a ConvNet's transferability given
  the target task. Although the optimal setting for some factors is similar for
  all tasks we considered, for other factors their optimal
  settings depend on the target task's distance from the source
  task. Table \ref{tab:tasks} shows the ordering of all tasks. There are a few exceptions to these general rules. For
  more detailed analysis refer to section
  \ref{sec:experiments}.}
\label{tab:best_practice} 
\end{table}
\renewcommand{\arraystretch}{1.0}

\subsubsection*{Outline of the paper} 
\label{sec:outline}
\begin{itemize}
\item We systematically identify and list the factors that
affect the transferability of ConvNet representation for visual
recognition tasks (Table \ref{tab:best_practice}, Section \ref{sec:experiments}). 
\item We provide exhaustive experimental evidence showing how these
factors should be set (Table \ref{tab:best_practice}, Section \ref{sec:experiments}).
\item We show these settings follow an interesting pattern which is
correlated with the distance between the source and target task,
(Figures 3-7 in Section \ref{sec:experiments}).
\item By optimizing the transferability factors we significantly improve (up to 50\% error reduction) state-of-the-art on 16 popular
visual recognition datasets (Table \ref{tab:final_results}) using a linear SVM for classification tasks and euclidean distance for instance retrieval.\\
\end{itemize} 

\subsubsection*{Related Works}
\label{sec:related_work}

\begin{table*}[]
\footnotesize
\centering
\begin{tabular}{l l l l l l}
 \multicolumn{6}{l}{$\,\,\,\,\,\,\,\,\,\xrightarrow{\makebox[1.5cm]{\textbf{Increasing distance from ImageNet}}}$}\\ 
\multirow{5}{*}{\rotatebox[origin=c]{90}{$\pmb{\xleftarrow{\makebox[1.3cm]{}}}$}}&&&&\\
&\multicolumn{1}{c}{\textcolor{red}{\textbf{Image Classification}}} & \multicolumn{1}{c}{\textcolor{Plum}{\textbf{Attribute Detection}}} & \multicolumn{1}{c}{\textcolor{blue}{\textbf{Fine-grained Recognition}}} & \multicolumn{1}{c}{\textcolor{SkyBlue}{\textbf{Compositional}}} & \multicolumn{1}{c}{\textcolor{YellowGreen}{\textbf{Instance Retrieval}}}\\
\cmidrule[1pt](l{2em}r{2em}){2-2}
\cmidrule[1pt](l{2em}r{2em}){3-3}
\cmidrule[1pt](l{2em}r{2em}){4-4}
\cmidrule[1pt](l{2em}r{2em}){5-5}
\cmidrule[1pt](l{2em}r{2em}){6-6}
&PASCAL VOC Object \cite{voc-2012}&H3D human attributes \cite{Bourdev11}&Cat\&Dog breeds \cite{Parkhi12}& VOC Human Action \cite{voc-2012}& Holiday scenes \cite{jegou08}\\
&MIT 67 Indoor Scenes \cite{Quattoni09}&Object attributes \cite{Farhadi09}&Bird subordinate \cite{Welinder2010}  & Stanford 40 Actions \cite{Yao11_ICCV} & Paris buildings \cite{Philbin08}\\
&SUN 397 Scene \cite{Xiao10}&SUN scene attributes \cite{Patterson12} & 102 Flowers \cite{Nilsback08} & Visual Phrases \cite{Sadeghi11} & Sculptures \cite{Arandjelovic11}\\
\\
\midrule
\end{tabular}
\caption{{\footnotesize Range of the 15 visual recognition tasks sorted categorically by their similarity to ILSVRC12 object image classification task.}}
\label{tab:tasks}
\end{table*}

The concept of learning from related tasks using neural networks and
ConvNets has appeared earlier in the literature
see \cite{Pratt92,Argyriou06,Gutstein08,Li10} for a few
examples. 
We describe two very recent papers which are the most relevant to our
findings in this paper.\\

In \cite{Agrawal:eccv:14} the authors investigate issues related to the training of ConvNets for the tasks of image
classification (SUN image classification dataset) and object detection
(PASCAL VOC 2007 \& 2012). The result of two of their investigations
are especially relevant to us. The first is that they show fine-tuning
a network, pre-trained with the ImageNet dataset, towards a target
task, image classification and object detection, has a positive effect
and this effect increases when more data is used for fine-tuning. They
also show that when training a network with ImageNet one should not
perform early stopping even if one intends to transfer the resulting
representation to a new task. These findings are consistent with a
subset of ours though our conclusions are supported by a larger and
wider set of experiments including more factors.\\
Yosinski et al. \cite{Yosinski:arxiv:14} show that the
transferability of a network trained to perform one source task to
solve another task is correlated with the distance between the source
and target tasks. Yosinski et al.'s source and target tasks are defined
as the classification of different subsets of the object categories in
ImageNet. Their definition of transferability comes from their
training set-up. First a ConvNet is trained to solve the source
task. Then the weights from the first $n$ layers of this source
network are \textit{transferred} to a new ConvNet that will be trained
to solve the target task. The rest of the target ConvNet's weights are
initialized randomly. Then the random weights are updated via
fine-tuning while the transferred weights are kept fixed. They show
that for larger $n$ the final target ConvNet, learned in this fashion,
performs worse and the drop in performance is bigger for the target
tasks most distant from the source task. This result corresponds to
our finding that the performance of the layer used for the ConvNet
representation is correlated to the distance between the source and
target task. Yosinki et al. also re-confirm that there are performance
gains to be made by fine-tuning a pre-trained network towards a target
task. However, our results are drawn from a wide range of
target tasks which are being used in the field of computer vision.
Furthermore, we have investigated more factors in addition to the
representation layer as listed in Table \ref{tab:best_practice}.



\pgfplotstableset{NetworkParams/.style ={%
        header=true,
        string type,
        font=\footnotesize,
        column type=l,
        every odd row/.style={
            before row=
        },
        every head row/.style={
            before row={%
              \topline\rowcolor{tableheadcolor}
             &  &            
            \multicolumn{4}{c}{\textbf{Convolutional layers}}
            & 
            \multicolumn{2}{c}{\textbf{FC layers}}
            \\
            \arrayrulecolor{tableheadcolor} \specialrule{6pt}{0pt}{-6pt} \arrayrulecolor{rulecolor}
            \cmidrule(r){3-6} \cmidrule(r){7-8} 
            \rowcolor{tableheadcolor}
            },
            after row={\midtopline}
        },
        every last row/.style={
            after row=\bottomline
        },
        columns/network/.style = {column name=\textbf{Network}},
        columns/TotalNum/.style = {column name=$N_T$},
        columns/ConvNum/.style = {column name=\#},
        columns/NumKernels/.style = {column name= $n_k$ per
          layer},
        columns/SzKernels/.style = {column name= kernel sizes per
          layer},
        columns/ConvOutDims/.style = {column name= output dimensions},
        columns/FCNum/.style = {column name=\#},
        columns/FCSize/.style = {column name=$n_h$ per layer},
        columns/OutputType/.style = {column name=function},
        columns/OutputNum/.style = {column name=$n_o$},
        col sep=&,
        row sep=\\
    }
}

\begin{table*}
  \caption{\textbf{Wider Networks:} Size details of the different ConvNet widths used in our experiments.}
  \label{tab:network_size}
\centering
{\centering
  \pgfplotstabletypeset[NetworkParams]{ 
    network & TotalNum & ConvNum & NumKernels & SzKernels & ConvOutDims & FCNum &
    FCSize 
    \\
    Tiny \;\;\;\;\;\,(A)&  14M & 5  & (24, 64, 96, 96, 64) & (11$\times$11, 5$\times$5,
    3$\times$3, 3$\times$3, 3$\times$3) &6$\times$6$\times$64& 3 & (4096, 1024, 1000) 
    \\
    \midrule
    Small \;\;\;\;(C)& 29M & 5 & (48, 128, 192, 192, 128) & (11$\times$11, 5$\times$5,
    3$\times$3, 3$\times$3, 3$\times$3) &6$\times$6$\times$128& 3 & (4096, 2048, 1000) 
    \\
    \midrule
    Medium (E)& 59M & 5 & (96, 256, 384, 384, 256) & (11$\times$11, 5$\times$5,
    3$\times$3, 3$\times$3, 3$\times$3) &6$\times$6$\times$256& 3 & (4096, 4096, 1000) 
    \\
    \midrule
    Large \;\;\;\;(F)& 138M & 6 & (96, 256, 512, 512, 1024, 1024) & (7$\times$7, 7$\times$7,
    3$\times$3, 3$\times$3, 3$\times$3, 3$\times$3) &5$\times$5$\times$1024& 3 & (4096, 4096, 1000) 
    \\
  }
}
\\[2pt]
\raggedright
{\captionfontsize{ 
The description of the notation in the table: $N_T$ is the
  total number of weights parameters in the network, $n_k$ is the
  number of kernels at a convolutional layer and $n_h$ is the number
  of nodes in a fully connected layer. For each network the output
  layer applies a soft max function and has 1000 output nodes. The
  networks are ordered w.r.t. their total number of parameters.}}
\label{tab:net_sizes_width}
\end{table*}

\section{Range of target tasks examined}
\label{sec:tasks}
To evaluate the transferability of the ConvNet representation we use a
wide range of 17 visual recognition tasks. The tasks are chosen from 5 different
subfields of visual recognition: object/scene image classification,
visual attribute detection, fine-grained classification, compositional
semantic recognition, instance retrieval (see
Table \ref{tab:tasks}). There are multiple ways one could order these
target tasks based on their similarity to the source task of object
image classification as defined by ILSVRC12.  Table \ref{tab:tasks}
gives our ordering and we now give the rationale for the ranking.

The group of tasks we consider furthest from the source task is
instance retrieval. Each task in this set has no explicit category
information and is solved by explicit matching to exemplars. While all
the other group of tasks involve classification problems and require
an explicit learning phases.

 We place attribute detection earlier than fine-grained recognition
 because these visual attributes\footnote{As an aside, depending on
   the definition of an attribute, the placement of an attribute
   detection task could be anywhere in the spectrum. For instance, one
   could define a fine-grained, local and compositional attribute
   which would then fall furthest from all other tasks (e.g. ``wearing
   glasses" in H3D dataset).} are usually the explanatory factors
 which separate the original object classes in ILSVRC and are thus
 expected to be naturally selected/highlighted by the ConvNet. Also,
 some attributes (e.g. four-legged) are defined as a superset of object
 classes (e.g. cat, dog, etc.). Another aspect of this pairwise
 ordering is that fine-grained recognition involves sometimes very
 subtle differences between members of a visual category. We suspect
 that a network trained for higher levels of object taxonomy (e.g.
 flowers in general) would not be sensitive to micro-scale visual
 elements necessary for fine-grained recognition.

Next comes perhaps the most interesting and challenging set of
category tasks -- the compositional recognition tasks. These tasks
include classes where the type of interactions between objects is the
key indicator and thus requires more sophistication to recognize than
the other category recognition tasks.

There are other elements which determine the closeness of a target task
to the source task. One is the distribution of the semantic classes
and images used within each category. For example the Pet
dataset \cite{Parkhi12} is the closest of the fine-grained tasks
because the ILSVRC classes include many different dog breeds. While,
sometimes the task just boils down to the co-occurrence of multiple
ILSVRC classes like the MIT indoor scenes. However, compositional recognition tasks
usually encode higher level semantic concepts to be inferred from the
object interactions, for instance a person holding violin is not
considered a positive sample for playing the violin in \cite{voc-2012}
nor is a person standing beside a horse considered as the action
``riding horse".

\section{Experiments}
\label{sec:experiments}
Now, we analyze the effect of each individual factor on the
transferability of the learnt representation. We divide the factors
into those which should be considered before learning a representation
(learning factors) and those which should be considered when using an
off-the-shelf network model (post-learning factors). 
\subsection{Learning Factors}
\vspace{0.15cm}
\subsubsection{Network Width}
\label{sec:network_size}

The ConvNet \texttt{AlexNet}\cite{Krizhevsky12}, the first very large
network successfully applied to the ImageNet challenge, has around 60
million parameters consisting of $\sim$5 million parameters in the
convolution layers and $\sim$55 million parameters in its fully
connected layers. Although this appears to be an unfeasibly large
parameter space the network was successfully trained using the
ImageNet dataset of 1.2 million images labelled with 1000 semantic
classes. More recently, networks larger than \texttt{AlexNet} have
been trained, in particular \overfeat\cite{Sermanet13}. Which of these
networks produces the best generic image representation and how
important is its size to its performance?\\

Here we examine the impact of the network's size (keeping its depth
fixed) on different tasks including the original ImageNet image-level
object classification. We trained 3 networks of different sizes using
the ILSVRC 2012 dataset and also included the \overfeatA network in
our experiments as the large network. Each network has roughly twice
as many parameters as we progress from the smallest to the largest
network. For all the networks we kept the number of units in the 6th
layer, the first fully connected layer, to 4096. It is this layer that
we use for the experiments where we directly compare networks. The
number of parameters is changed mainly by halving the number of
kernels and the number of fully connected neurons (except the fixed
one).

\begin{figure}[t]
\centering
    \includegraphics[width=1\linewidth]{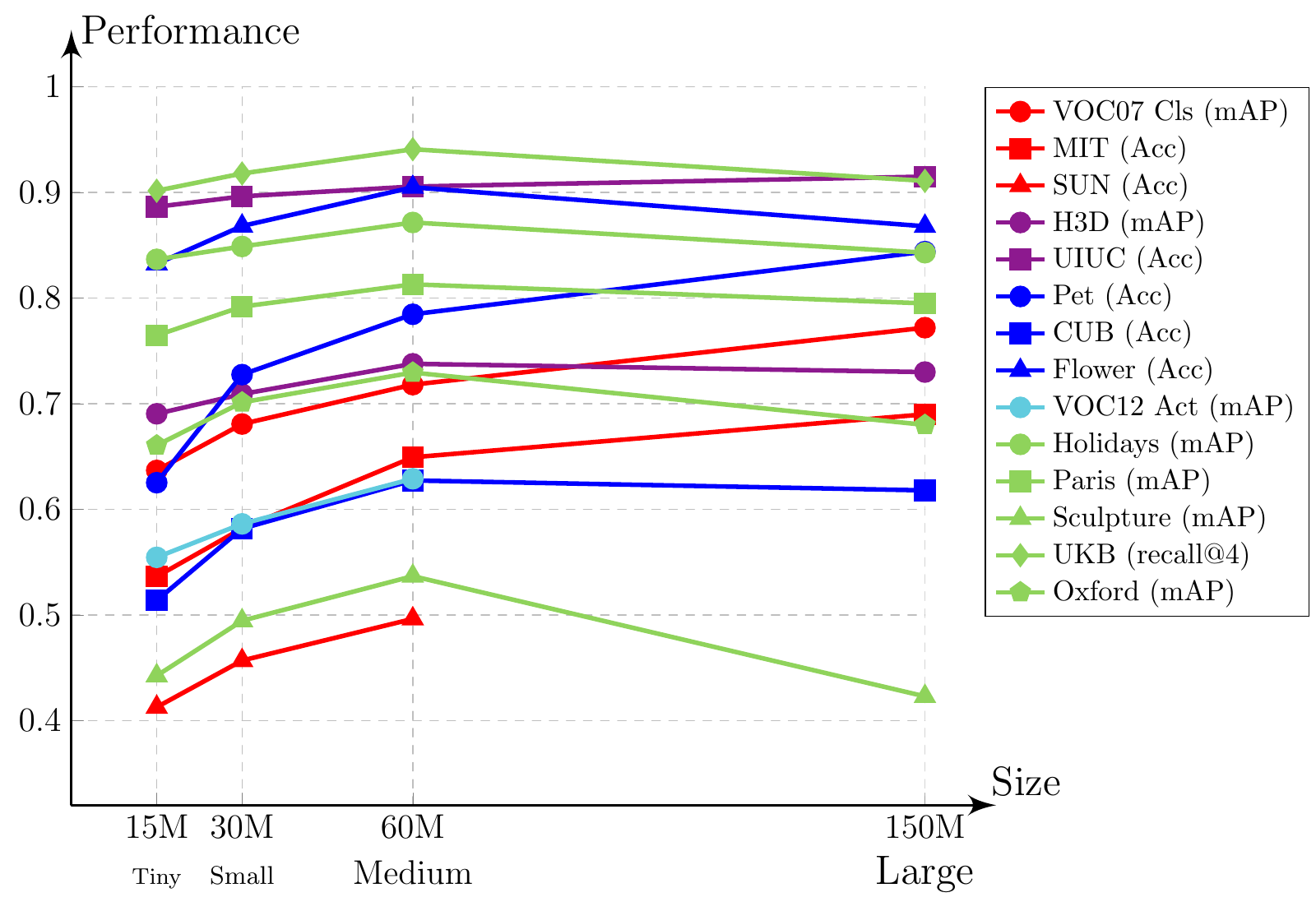}%
\caption{\captionfontsize \textbf{Network Width:} Over-parametrized networks (\overfeat) can be
  effective when the target task is close to the labelled
  data. However, the performance on more distant tasks can suffer from
  over-specialization when the number of network parameters is
  increased. But overall under-parametrized networks (Tiny) are unable to
  generalize as well. Since the Tiny network has 10 times fewer
  parameters than \overfeatA while preserving most of the performance,
  it could be useful for scenarios where real-time computation is an
  issue.}%
\label{fig:network_size}
\end{figure}

Figure \ref{fig:network_size} displays the effect of changing the
network size on different visual recognition tasks/datasets.  The
largest network works best for Pascal VOC object image classification,
MIT 67 indoor scene image classification, UIUC
object attribute, and Oxford pets dataset. 
On the other hand, for all the retrieval tasks the performance of the
over-parametrized \overfeatA network consistently suffers because it
appears the generality of its representation is less than those of the
smaller networks. Another interesting observation is that, if the
computational efficiency at test time is critical, one can decrease
the number of network parameters by orders of 2 (Small or Tiny
network) for different tasks but the degradation of the final
performance is sublinear in some cases.\\

\subsubsection{Network Depth}
\vspace{0.15cm}
\begin{figure}[t!]
\centering
    \includegraphics[width=1\linewidth]{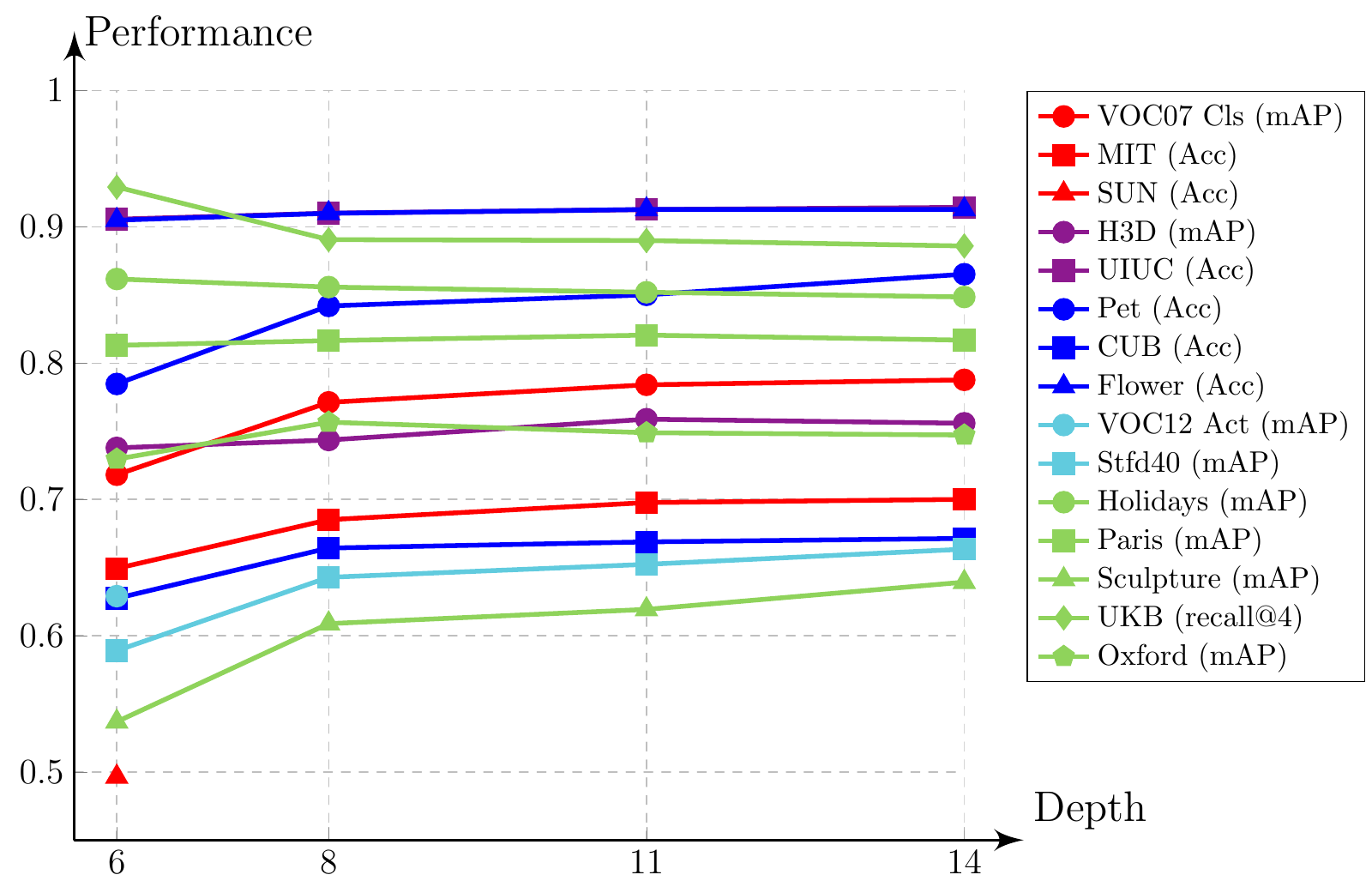}
\caption{\captionfontsize \textbf{Network Depth:} Over-parametrizing networks by the number of convolutional layers is
  effective for nearly all the target tasks. While a saturation can be observed in some tasks there is no significant performance drop as opposed to the trend observed when 
  over-parametrizing by increasing its width. The number on the x-axis indicates the number of convolutional layers of the network. The representation is taken from the first fully connected layer right after the last convolutional layer.}%
\label{fig:network_depth}
\end{figure}

\begin{figure*}[t!]
\centering
\begin{subfigure}[b]{0.73\linewidth}
    \includegraphics[width=1\linewidth]{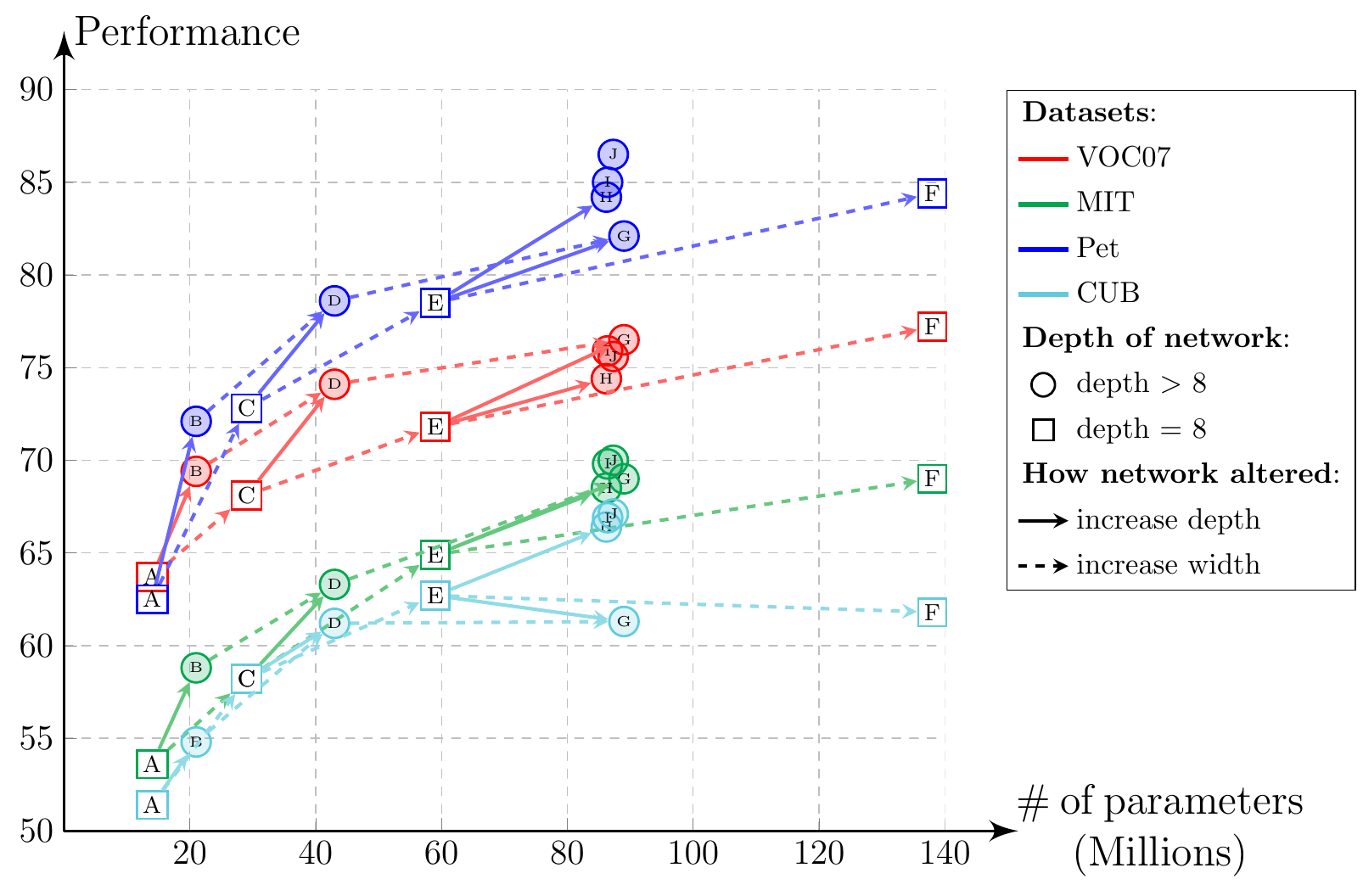}
\end{subfigure}
\quad
\begin{subfigure}[b]{0.24\linewidth}
    \includegraphics[width=1\linewidth]{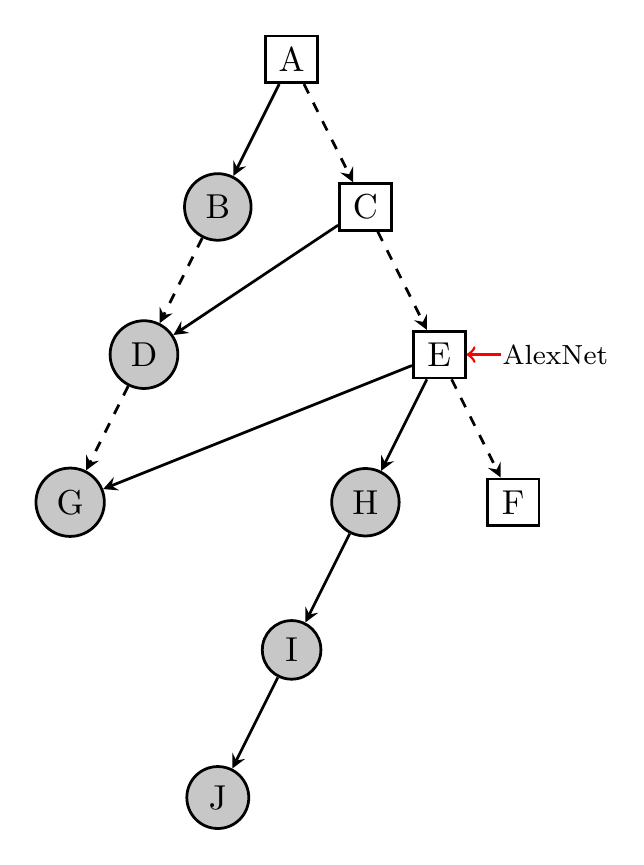}
    \vspace{2cm}

\end{subfigure}
\caption{\captionfontsize \textbf{Depth versus Width:} Over-parametrization of networks can be done by increasing either of its width and depth or by both. In this figure, effect of increasing depth is illustrated on the final performance and can be compared with that of increasing width. Solid lines indicate the evolution of performance when depth of the network is increased. Circles indicates networks of depth 8 (\emph{e.g.} AlexNet) and squares are used for deeper networks. Dashed lines, on the other hand, declares an increase in the width of the network. It can be seen from the results that increasing the depth is more efficient in number of parameters per unit of performance gained in various datasets (solid lines have higher slopes). Refer to Tables \ref{tab:net_sizes_depth} and \ref{tab:net_sizes_width} for exact architecture of the networks used in this experiment. The representation is taken from the first fully connected layer right after the last convolutional layer. The tree on the right, depicts the relationship between different networks.} %
\label{fig:width_vs_depth}
\end{figure*}

Increasing the network width (number of parameters at each layer) is
not the only way of over-parameterizing a ConvNet. In fact,
\cite{Szegedy14, Simonyan14} have shown that deeper convolutional
networks with more layers achieve better performance on the ILSVRC14
challenge. In a similar spirit, we over-parametrize the network by
increasing the number of convolutional layers before the fully
connected layer from which we extract the representation. Figure
\ref{fig:network_depth} shows the results by incrementally increasing
the number of convolutional layers from 5 to 13 (the architectures of these networks is described in Table \ref{tab:net_sizes_depth}). As this number is
increased, the performance on nearly all the datasets increases.

 The only tasks for which the results degrade are the retrieval tasks
 of UKB and Holidays. Interestingly, these two tasks involve measuring
 the visual similarity between specific instances of classes strongly
 presented in ImageNet (e.g. a specific book, bottle or musical
 instrument in UKB, and wine bottle, Japanese food in Holidays
 dataset). It is, thus, expected that the representation becomes more
 invariant to instance level differences as we increase the complexity
 of the representation with more layers.

If we compare the effect of increasing network depth to network width
on the final representation's performance, we clearly see that
increasing depth is a much more stable over-parametrization of the
network. Both increasing width and depth improve the performance on
tasks close to the source task. However, increasing the width seems to
harm the transferability of features to distant target tasks more than
increasing depth does. This could be attributed to the fact that
increasing depth is more efficient than increasing width in terms of the number of
parameters for representing more complex patterns, next section studies this in a separate experiment. Finally, more
layers means more sequential processing which hurts the
parallelization. We have observed the computational complexity for
learning and using deep ConvNets increases super-linearly with the
number of layers. So, learning a very wide network is computationally
cheaper than learning a very deep network. These issues means the
practitioner must decide on the trade-off between training speed and
performance.\\

\subsubsection{Width vs Depth}
\vspace{0.15cm}
It is more indicative to directly compare the effect of increasing width and depth on generality of the learned representation. For that purpose we train deep networks of various depth. Particularly, we train a network of depth 16 with similar width to the Tiny, Small and Medium networks in the previous section. Table \ref{tab:net_sizes_depth} lists the deep networks and their architectures. We compare the results of 10 different networks on four target tasks in Figure \ref{fig:width_vs_depth}. By connecting different networks with solid (dashed) directed edges we show the performance of the a deeper (wider) network. It can be observed that increasing the parameters of the network by increasing its depth is a more efficient over-parametrization than increasing its width. That is the slope of solid segments are consistently higher.\\
It should be noted that most of the parameters of a convolutional network usually is at the first fully connected layer. Thus, the number of outputs of the last convolutional layer (which depends on the preceding subsampling layers) is the major factor for the network size. For example, going from network H to I and then to J slightly increases the number of parameters but considerably increases the performance on the target tasks.\\
Training the deeper networks is tricky and needs to be done in various stages. For networks with convolutional layers of more than 5 (Medium) we increased the number of convolutional layers by 3 at each time. Then trained the network for a few epochs ($<10$) with fixed learning rate ($0.01$) and initialized the first convolutional layers of the next network with those of the shallower network. The new convolutional layers and fully connected layers were initialized using random gaussian noise.\\



\pgfplotstableset{NetworkParams/.style ={%
        header=true,
        string type,
        font=\footnotesize,
        column type=l,
        every odd row/.style={
            before row=
        },
        every head row/.style={
            before row={%
              \topline\rowcolor{tableheadcolor}
             &  &            
            \multicolumn{3}{c}{\textbf{Convolutional layers}}
            & 
            \multicolumn{2}{c}{\textbf{FC layers}}
            \\
            \arrayrulecolor{tableheadcolor} \specialrule{6pt}{0pt}{-6pt} \arrayrulecolor{rulecolor}
            \cmidrule(r){3-5} \cmidrule(r){6-7} 
            \rowcolor{tableheadcolor}
            },
            after row={\midtopline}
        },
        every last row/.style={
            after row=\bottomline
        },
        columns/network/.style = {column name=\textbf{Network}},
        columns/TotalNum/.style = {column name=$N_T$},
        columns/ConvNum/.style = {column name=\#},
        columns/NumKernels/.style = {column name= $n_l\times n_k$ per
          layer},
        columns/SzKernels/.style = {column name= kernel sizes per
          layer},
        columns/ConvOutDims/.style = {column name= out dim},
        columns/FCNum/.style = {column name=\#},
        columns/FCSize/.style = {column name=$n_h$ per layer},
        columns/OutputType/.style = {column name=function},
        columns/OutputNum/.style = {column name=$n_o$},
        col sep=&,
        row sep=\\
    }
}
\pgfplotstableset{NetworkParams2/.style ={%
        header=true,
        string type,
        font=\footnotesize,
        column type=l,
        every odd row/.style={
            before row=
        },
        every head row/.style={
            before row={%
              \topline\rowcolor{tableheadcolor}
             &  &
            \multicolumn{2}{c}{\textbf{Convolutional layers}}
            & 
            \multicolumn{2}{c}{\textbf{FC layers}}
            \\
            \arrayrulecolor{tableheadcolor} \specialrule{6pt}{0pt}{-6pt} \arrayrulecolor{rulecolor}
            \cmidrule(r){3-4} \cmidrule(r){5-6} 
            \rowcolor{tableheadcolor}
            },
            after row={\midtopline}
        },
        every last row/.style={
            after row=\bottomline
        },
        columns/network/.style = {column name=\textbf{Network}},
        columns/TotalNum/.style = {column name=$N_T$},
        columns/ConvNum/.style = {column name=\#},
        columns/NumKernels/.style = {column name= $n_l\times n_k$},
        columns/SzKernels/.style = {column name= size},
        columns/ConvOutDims/.style = {column name= output dim},
        columns/FCNum/.style = {column name=\#},
        columns/FCSize/.style = {column name=$n_h$ per layer},
        columns/OutputType/.style = {column name=function},
        columns/OutputNum/.style = {column name=$n_o$},
        col sep=&,
        row sep=\\
    }
}
\begin{table*}
\setlength{\tabcolsep}{2.5pt}
  \caption{\textbf{Deeper Networks:} Size details of the different deep ConvNets used in our experiments.}
  \label{tab:network_size}
\centering
{\centering
  \pgfplotstabletypeset[NetworkParams]{ 
    network & TotalNum & ConvNum & NumKernels & ConvOutDims & FCNum &
    FCSize 
    \\
    Deep8   & 85M & 5  & (1$\times$64, 1$\times$128, 3$\times$256) &8$\times$8$\times$256& 3 & (4096, 4096, 1000) 
    \\
    \midrule
    Deep11 (H) & 86M & 8 &  (1$\times$64, 3$\times$128, 4$\times$256) &8$\times$8$\times$256& 3 & (4096, 4096, 1000) 
    \\
    \midrule
    Deep13 (I) & 86M & 10 & (2$\times$64, 4$\times$128, 4$\times$256) &8$\times$8$\times$256& 3 & (4096, 4096, 1000) 
    \\
    \midrule
    Deep16 (J) & 87M & 13 & (2$\times$64, 5$\times$128, 6$\times$256) &8$\times$8$\times$256& 3 & (4096, 4096, 1000) 
    \\
  }
}
{
  \pgfplotstabletypeset[NetworkParams2]{ 
    network & TotalNum & NumKernels & ConvOutDims & FCNum &
    FCSize 
    \\
    Deep Tiny\;\;\;(B)& 21M & 13$\times$64 & 8$\times$8$\times$64 & 3 & (4096, 4096, 1000)     \\
    \midrule
    Deep Small (D) & 43M &13$\times$128& 8$\times$8$\times$128& 3 & (4096, 4096, 1000)     \\
    \midrule
    Deep Med\;\;\;(G) & 89M &13$\times$256& 8$\times$8$\times$256& 3 & (4096, 4096, 1000)     \\
    \\
    \\
  }
}
\\[2pt]
\raggedright
{\captionfontsize{ 
The description of the notation in the table: $N_T$ is the
  total number of weights parameters in the network, $n_k$ is the
  number of kernels at a convolutional layer, $n_l$ is the number of layers with $n_k$ kernels, and $n_h$ is the number
  of nodes in a fully connected layer. All the kernels have spatial size of 3x3. For each network the output
  layer applies a SoftMax function and has 1000 output nodes. The
  networks are ordered w.r.t. their total number of parameters. Note that these networks are re-trained for our experiments and the models differ from that of \cite{Simonyan14}, for instance we do not use multi-scale input and our input image size is 227x227, we do random cropping as implemented in \texttt{Caffe}, etc.}}
\label{tab:net_sizes_depth}
\end{table*}

\subsubsection{Early Stopping}
\begin{figure}[t!]
  \centering
  \includegraphics[width=1\linewidth]{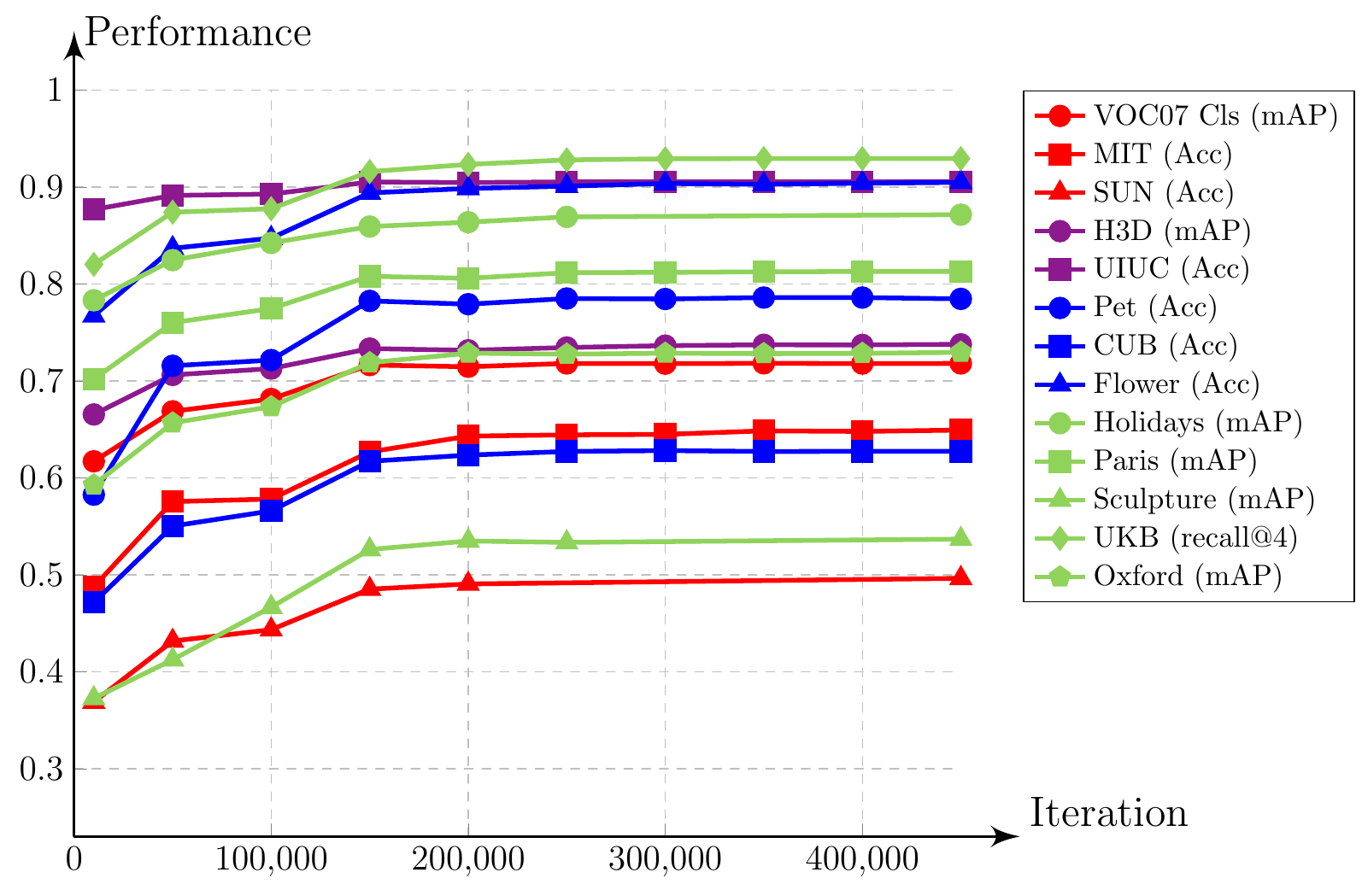}
  \caption{\captionfontsize \textbf{Early Stopping:} Plotted above is the
    performance of the representation extracted from layer 6 of the
    \alexnetA ConvNet versus the number of iterations of SGD used to
    train the initial network. It can be seen that early stopping,
    which can act as a regularizer, does not help to produce a more
    transferable representation.}
\label{fig:iter_effect}
\end{figure}

\vspace{0.15cm}
\label{sec:training_iterations}
Early stopping is used as a way of controlling the generalization of a
model. It is enforced by stopping the learning before it has converged
to a local minima as measured by monitoring the validation loss. This
approach has also been used to improve the generalization of
over-parametrized networks \cite{Bengio12}. It is plausible to expect
that the transferability increases with generalization. Therefore, we
investigate the effect of early stopping on the transferability of
learnt representation. Figure \ref{fig:iter_effect} shows the
evolution of the performance for various target tasks at different
training iterations. The performance of all tasks saturates at 200K
iterations for all the layers and even earlier for some
tasks. Surprisingly, it can be seen that early stopping does not
improve the transferability of the features whatsoever. However, in this experiments the training does not show strong symptoms of over-fitting. We have observed that if training of the source ConvNet exhibits overfitting (such as in fine-tuning with landmark dataset for improved performance on instance retrieval) early stopping can help to learn more transferable features.\\

\subsubsection{Source Task}
\vspace{0.15cm}
\begin{figure}[thpb]
   \centering      
   \includegraphics[width=1\linewidth]{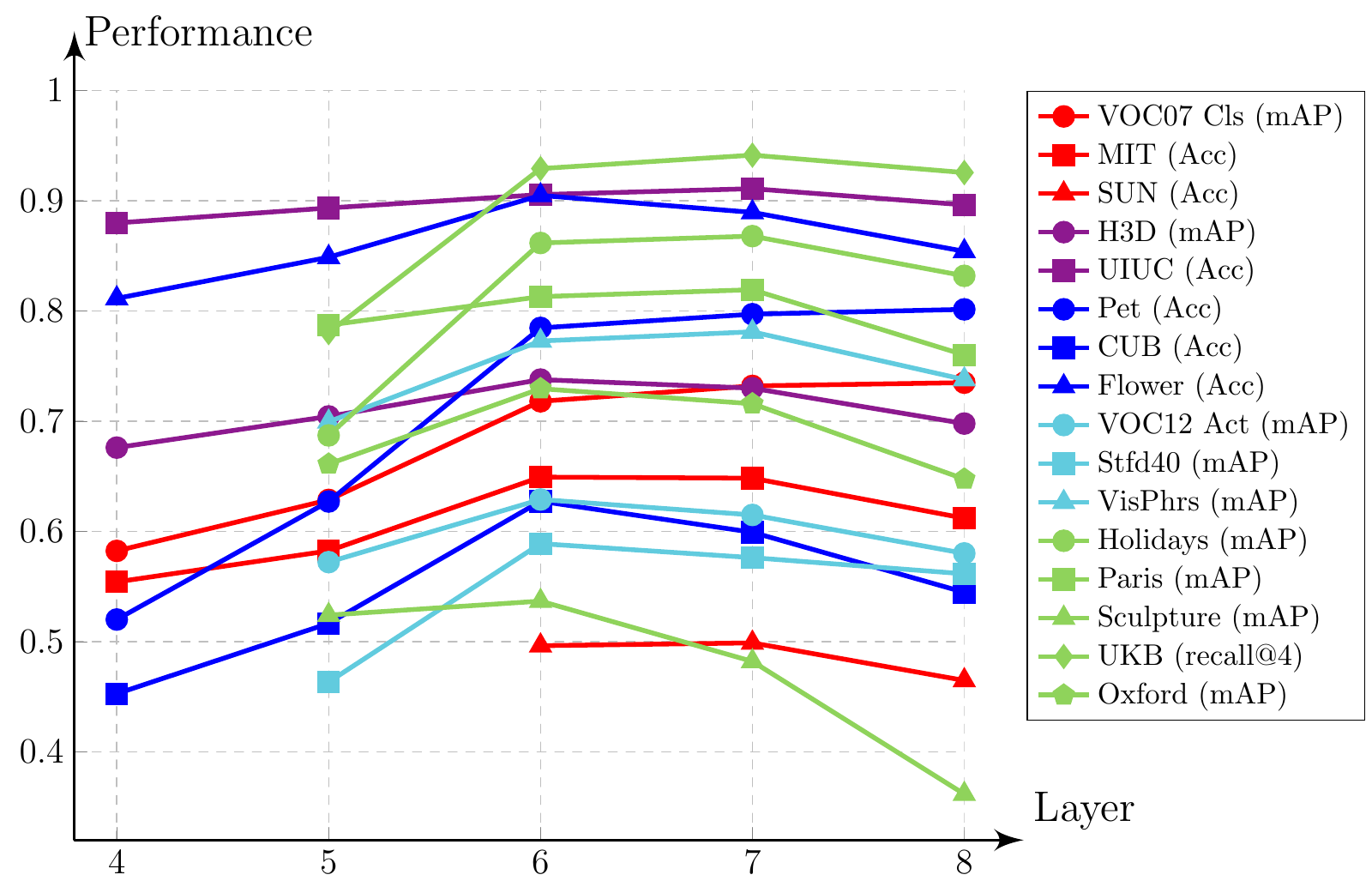}
\caption{\captionfontsize \textbf{Representation Layer:} Efficacy of representations extracted from \alexnet's different layers for different
visual recognition tasks. A distinct pattern can be observed: the
further the task moves from object image classification, the earlier
layers are more effective. For instance, layer 8 works best for PASCAL VOC image classification which is very similar to ImageNet while the best performance for all retrieval tasks is at layer 6.}
\label{fig:layer_effect}
\end{figure}

It is natural to expect that one of the most important factors for a learnt
representation to be generic is the properties of the source task (disregarding the number of images in the source dataset). The recent development of another
large scale dataset called the Places Dataset \cite{Zhou14} labelled
with scene classes enabled us to analyze this factor. Table
\ref{tab:source_tasks} shows the results for different source tasks of
ImageNet, Places, and a hybrid network. The hybrid network is made by
combining the ImageNet images with those of the Places dataset. The
label set is increased accordingly \cite{Zhou14}. It can be observed
that results for the tasks very close to the source tasks are improved
with the corresponding models (MIT, SUN for Places network). Another
interesting observation is that ImageNet features seem to achieve a
higher level of generalization for further away tasks. One explanation
is that the set of labels is more diverse. Since the number of images
in ImageNet is smaller, it shows the importance of diversity of labels
as opposed to the number of annotated images when the objective is to
achieve a more transferable representation. More concrete experiments on this phenomenon is conducted in the next section.

The Hybrid model boosts the transferability of the Places network but
still falls behind the ImageNet network for more distant tasks. This
could be again due to the fact that the number of images from the
Places dataset dominates those of the ImageNet dataset in training the
Hybrid model and as a consequence it is more biased toward the Places
Network. 

In order to avoid this bias, in another experiment, we
combined the features obtained from the ImageNet network and the
Places network as opposed to Hybrid network, and interestingly this
late fusion works better than Hybrid model (the Hybrid model
where the number of dimensions of the representation is increased to
8192 works worse \cite{Chatfield14}). In fact, it achieves the best results on all tasks except for subcategory recognition tasks for which scenes are irrelevant and probably just add noise to the descriptors. \\

\begin{table*}[]
\setlength{\tabcolsep}{7.5pt}
  \centering
    \centering
    \footnotesize
      \begin{tabular}{l c l l l c l l l c l l l}
        &\multicolumn{3}{c}{\textcolor{red}{\textbf{Image Classification}}} & \multicolumn{2}{c}{\textcolor{Plum}{\textbf{Attribute Detection}}} & \multicolumn{3}{c}{\textcolor{blue}{\textbf{Fine-grained Recognition}}} & \multicolumn{1}{c}{\textcolor{SkyBlue}{\textbf{Compositional}}} & \multicolumn{3}{c}{\textcolor{YellowGreen}{\textbf{Instance Retrieval}}}\\
        \cmidrule[1pt](l{2em}r{2em}){2-4}
        \cmidrule[1pt](l{2em}r{2em}){5-6}
        \cmidrule[1pt](l{2em}r{2em}){7-9}
        \cmidrule[1pt](l{2em}r{2em}){10-10}
        \cmidrule[1pt](l{2em}r{2em}){11-13}
        \textbf  {Source task} 	&VOC07			&MIT			&SUN			&H3D			&UIUC			&Pet			&CUB			&Flower			&Stanf. Act40			&Oxf.			&Scul.			&UKB\\ 
        \midrule ImageNet 		&71.6			&64.9			&49.6			&73.8			&\textbf{90.4}	&\textbf{78.4}	&\textbf{62.7}	&\textbf{90.5}	&58.9			&71.2			&52.0			&93.0\\ 
        \midrule Places			&68.5			&69.3			&55.7			&68.0			&88.8			&49.9			&42.2			&82.4			&53.0			&70.0			&44.2			&88.7\\
        \midrule Hybrid 		&72.7			&69.6			&56.0			&72.6			&90.2			&72.4			&58.3			&89.4			&58.2			&\textbf{72.3}	&52.3			&92.2\\
        \midrule Concat 		&\textbf{73.8}	&\textbf{70.8}	&\textbf{56.2}	&\textbf{74.2}	&\textbf{90.4}	&75.6			&60.3			&90.2		&\textbf{59.6}			&72.1			&\textbf{54.0}	&\textbf{93.2}\\
        \bottomline
      \end{tabular}
  \caption{\captionfontsize \textbf{Source Task:} Results on all tasks using representations optimized for different source tasks. ImageNet is the representation used for all experiments of this paper. Places is a new ConvNet trained on 3.5M images labeled with scene categories \cite{Zhou14}. Hybrid is a model proposed by Zhou et al. \cite{Zhou14} which combines the ImageNet and Places datasets and train a single network for the combination. Concat indicates results of concatenating the feature obtained from ImageNet ConvNet and Places ConvNet for each input image. All results are for first fully connected layer (FC6).}
\label{tab:source_tasks}
\end{table*}

\subsubsection{Diversity and Density of Training Data}
\begin{figure*}
        \centering
        \begin{subfigure}[b]{0.5\linewidth}
   \includegraphics[clip=true, trim=0 0 0 5,width=1\linewidth]{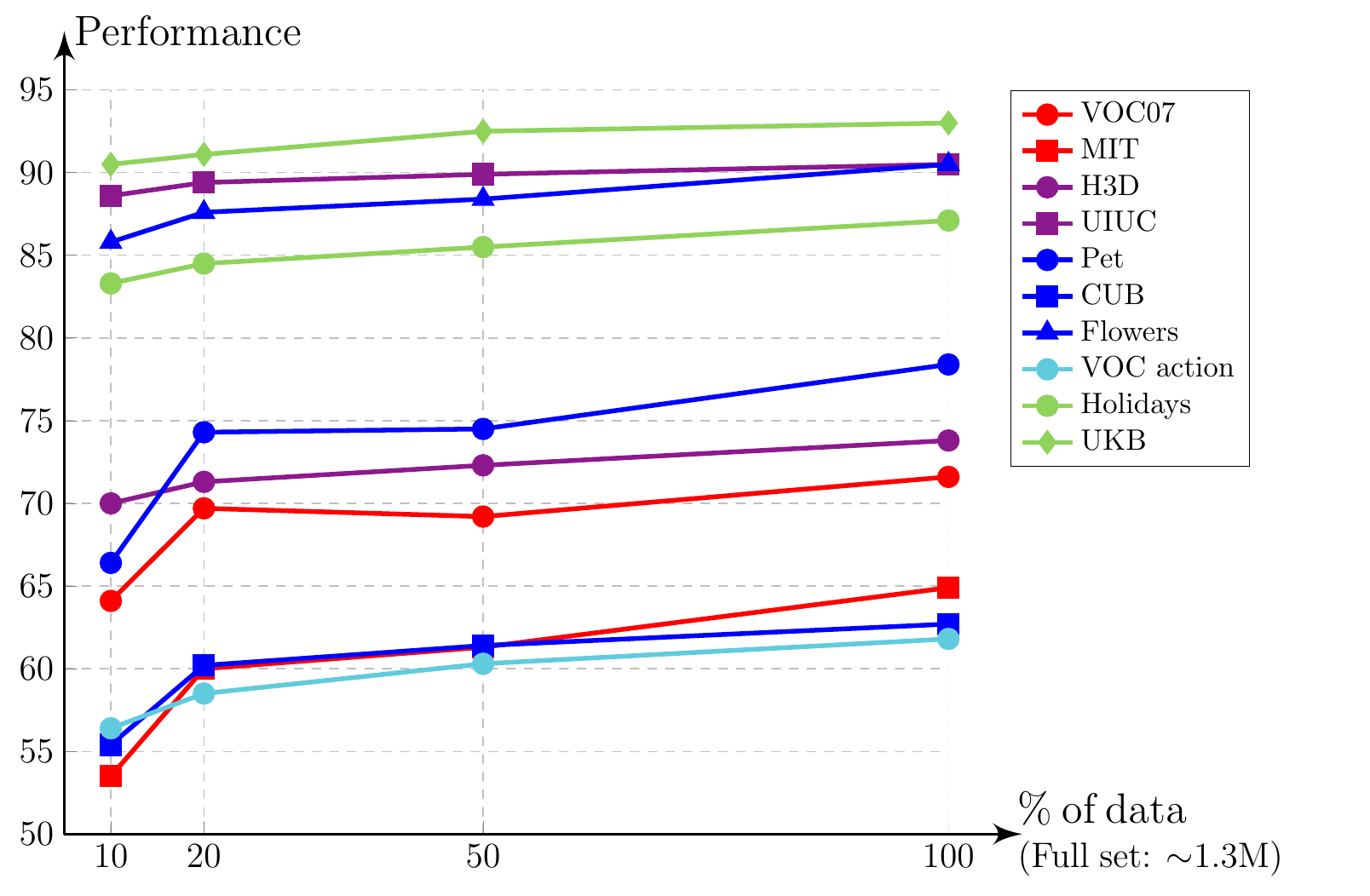}
\caption{\captionfontsize \textbf{Increasing Density:} We trained 3 "Medium" networks using different number of images per ILSVRC 2012 class. We decreased the number of images for each class to 10\%, 20\% and 50\% of its original set separately. The images are sampled randomly. Heavier regularization is applied for networks with smaller data size.}
\label{fig:increase_data_density}
        \end{subfigure}
        \begin{subfigure}[b]{0.45\linewidth}
   \includegraphics[clip=true, trim=0 9 0 0,width=1\linewidth]{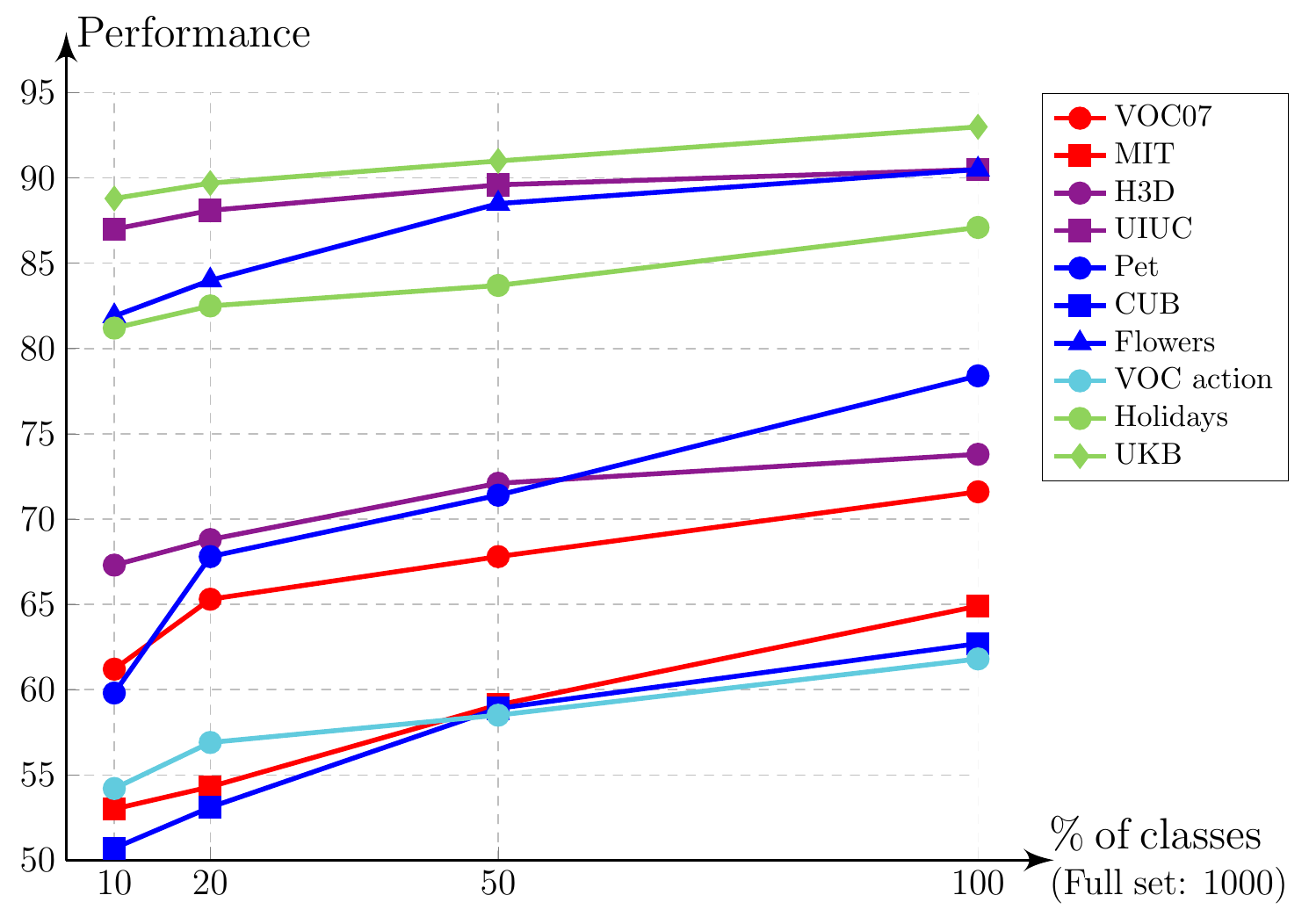}
\caption{\captionfontsize \textbf{Increasing Diversity:} We trained 3 "Medium" networks using different number classes from ImageNet. We decreased the number of classes to 100, 200 and 500 but kept the images within each class the same as ILSVRC 2012. The classes are sampled randomly. Heavier regularization is applied for networks with smaller data size.}
\label{fig:increase_data_diversity}
        \end{subfigure}%
\caption{\captionfontsize \textbf{Density versus Diversity of Training Data:} Changing the size of training data by altering number of images per class versus number of classes leads to different performances on transferred tasks. Interestingly, the results using lower diversity is consistently inferior to the performances obtained by lower density (in a point-to-point comparison). That means diversity of the training data is more important than density of training data when transferring the learnt representation to various tasks. This trend is observed regardless of the distance of the target task to the original task.}
\label{fig:density_vs_diversity}
\end{figure*}

\vspace{0.15cm}
We saw in the previous section that the distribution of training data for the source task affects the transferability of the learned representation. Annotating millions of images with various labels used for learning a generic representation is expensive and time-consuming. Thus, choosing how many images to label and what set of labels to include is a crucial question. In addition to the source tasks in the previous section, we now examine the influence of statistical properties of the training data such as density and diversity of the images. In this experiment we assume that ImageNet classes indicate different modes of the training data distribution. Such that, by increasing the number of images per class we control the density of the distribution. Moreover, the diversity of the training data can be increased by including additional classes. \\
In order to compare the effect of diversity and density of training data on the transferability of the learned representation we assume a situtation where there is a certain budget for the number of images to be annotated. Particularly, we experiment for the training data size of 10\%, 20\%, 50\% of the ILSVRC12 1.3 million images. Then the dataset is either constructed using stratified sampling from \textit{all classes} (reduced density with the same diversity) or by random sampling of the classes with \textit{all of their samples} (reduced diversity with the same density).\\
Figure \ref{fig:density_vs_diversity} plots the results for various tasks when decreasing density (Figure \ref{fig:increase_data_density}) or diversity (Figure \ref{fig:increase_data_diversity}) of the source training set from that of ILSVRC12. It can be seen that increasing both diversity and density consistently helps all the tasks and there is still no indication of saturation at the full set of 1.3 million images. Thus there is still room for annotating more images beyond ILSVRC and that most probably increases the performance of the learned representation on all the target tasks. Furthermore no clear correlation can be observed between the degradation of the performances and the distance of the target task. Most importantly, decreasing the diversity of the source task seems to hurt the performance on the target tasks more significantly (slopes on the right plot are higher than left plot). In fact, a point to point comparison of the two plots reveals that when having certain budget of images, increasing diversity is crucially more effective than density. This could be because of the higher levels of feature sharing which happens at higher diversities which helps the generalization of the learned representation and thus is more beneficial for transfer learning.\\
The network architecture used for this experiment is Medium (AlexNet), so the number of parameters remains the same for all of the experiments. However training the network on the smaller datasets needed a heavier regularization by increasing the weight decay and dropout ratio at the fully connected layers. Without heavy regularization training a Medium network on 10\% or 20\% exhibits signs of over-fitting in the early stages of training.\\

\subsection{Post-learning Factors}
\vspace{0.3cm}
\subsubsection{Network Layer}
\vspace{0.15cm}
\label{sec:network_layers}
Different layers of a ConvNet potentially encode different levels of
abstraction. The first convolutional layer is usually a collection of
Gabor like gray-scale and RGB filters. On the other hand the output
layer is directly activated by the semantic labels used for
training. It is expected that the intermediate layers span the levels
of abstraction between these two extremes. Therefore, we used the
output of different layers as the representation for our tasks'
training/testing procedures. The performance of different layers of
the pre-trained ConvNet (size: Medium) on ImageNet is shown in
figure \ref{fig:layer_effect} for multiple tasks.

We observe the same pattern as for the effect of network width. The last
layer (1000-way output) is only effective for the PASCAL VOC
classification task. In the VOC task the semantic labels are a subset
of those in ILSVRC12, the same is true for the Pet dataset
classes. The second fully connected layer (Layer 7) is most effective
for the UIUC attributes (disjoint groups of ILSVRC12),and MIT indoor
scenes (simple composition of ILSVRC12 classes). The first fully
connected layer (Layer 6) works best for the rest of the datasets
which have semantic labels further away from those used for optimizing
the ConvNet representation. An interesting observation is that the
first fully connected layer demonstrates a good trade-off when the
final task is unknown and thus is the most generic layer within the
scope of our tasks/datasets.

Although the last layer units act as probabilities for ImageNet classes, note that results using the last layer with 1000 outputs are surprisingly
effective for almost all the tasks. This shows that a high order of image-level information lingers even to the very last layer of the network. It should be mentioned that obtaining results of instance retrieval on convolutional layers is computationally prohibitive and thus they are not included. However, in a simplified scenario, the retrieval results showed a drastic decrease from layer 6 to 5.

\begin{figure}[t!]
   \centering      
   \includegraphics[width=1.04\linewidth]{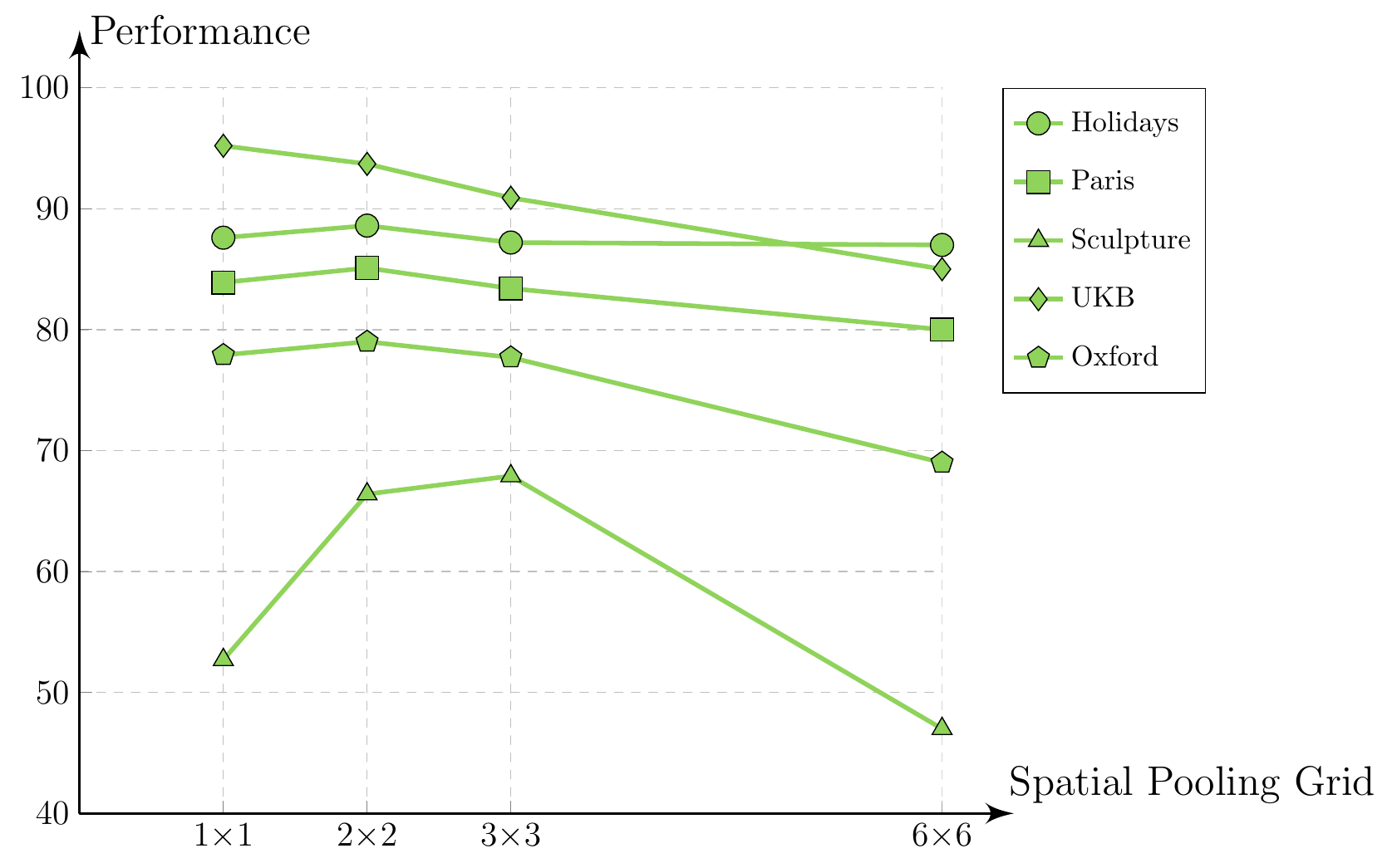}
\caption{\captionfontsize \textbf{Spatial Pooling:} In order to obtain meaningful results for retrieval with representations from convolutional layers we applied spatial pooling of different sizes for different tasks. Objects of more complex structures such as sculptures and buildings need more spatial resolution for optimal performance.}
\label{fig:pooling}
\end{figure}


\subsubsection{Spatial Pooling}
In the last subsection, we observed that the best representation for
retrieval tasks is the first fully connected layer by a significant
margin. We further examined using the last convolutional layer in its
original form as the representation for retrieval in a simplified
scenario but achieved relatively poor results. In order to make the
convolutional layer suitable, in this experiment, spatial pooling is
applied to the last convolutional layers output. We use max pooling in this experiment. An spatial pooling with a grid of $1\times1$ is equivalent to a soft bag of words representation over the whole image, where words are convolutional kernels. Figure \ref{fig:pooling} shows the results of different pooling grids for all the retrieval tasks. For the retrieval tasks, where the shapes are
more complicated like sculptures and historical buildings, a higher
resolution of pooling is necessary. \\

\begin{figure*}[h]
        \centering
        \begin{subfigure}[t!]{0.45\linewidth}
        \vspace{0.cm}
   \includegraphics[width=1\linewidth]{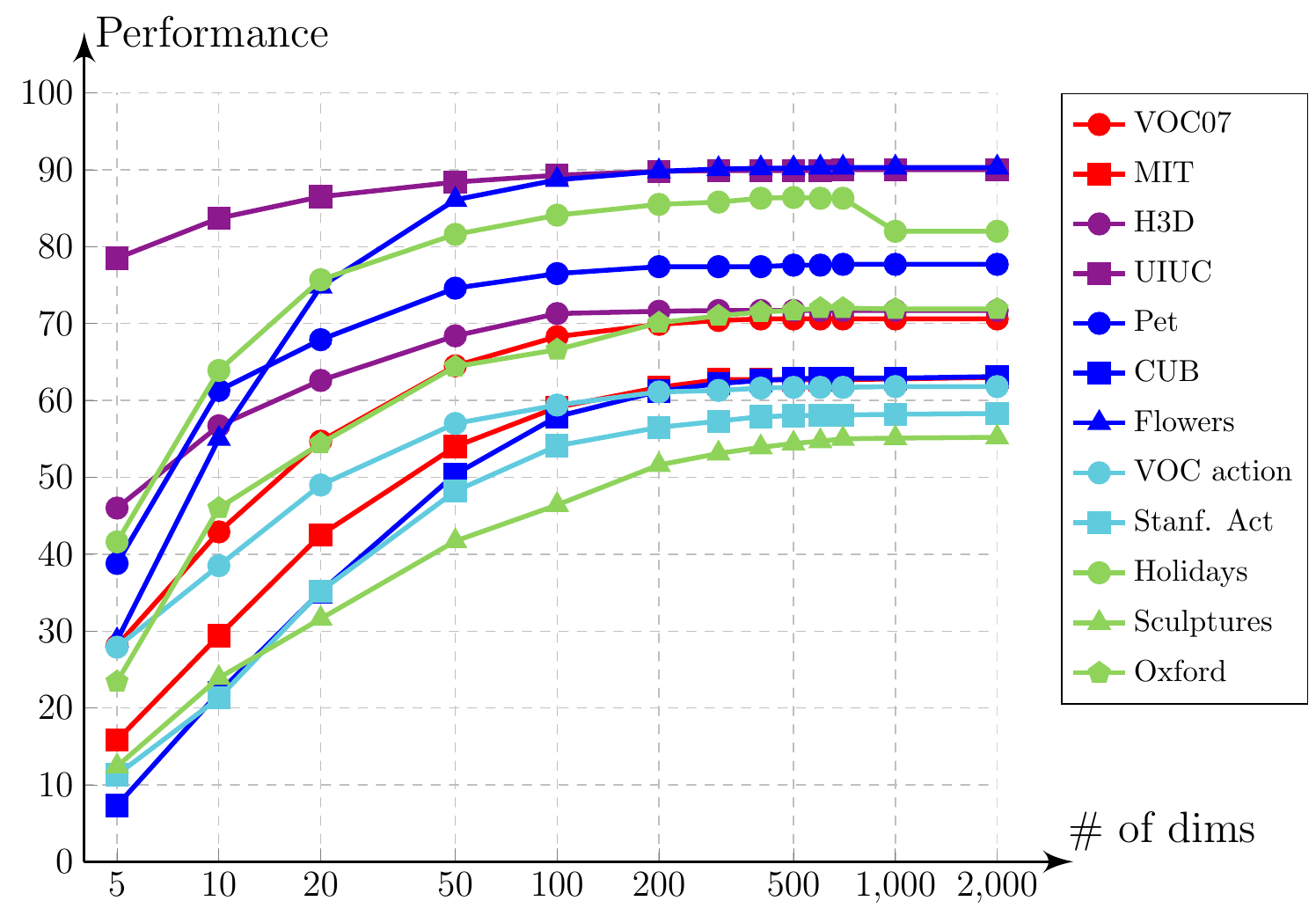}
\caption{\captionfontsize \textbf{Effective Dimensionality:} It can be observed that almost all tasks reach their maximum performance at an dimensionality of below 500 indicating a low (class-conditional) effective dimensionality of ConvNet representations. The accuracy of all tasks for dimensions under 50 are surprisingly high. Thus, he fact that these transformations are obtained using a linear transform supports capability of ConvNet in generalization by disentangling underlying generating factors.}
\label{fig:pca}
        \end{subfigure}%
        \qquad
        \begin{subfigure}[t!]{0.5\linewidth}
        \vspace{0.15cm}
   \includegraphics[width=0.5\linewidth]{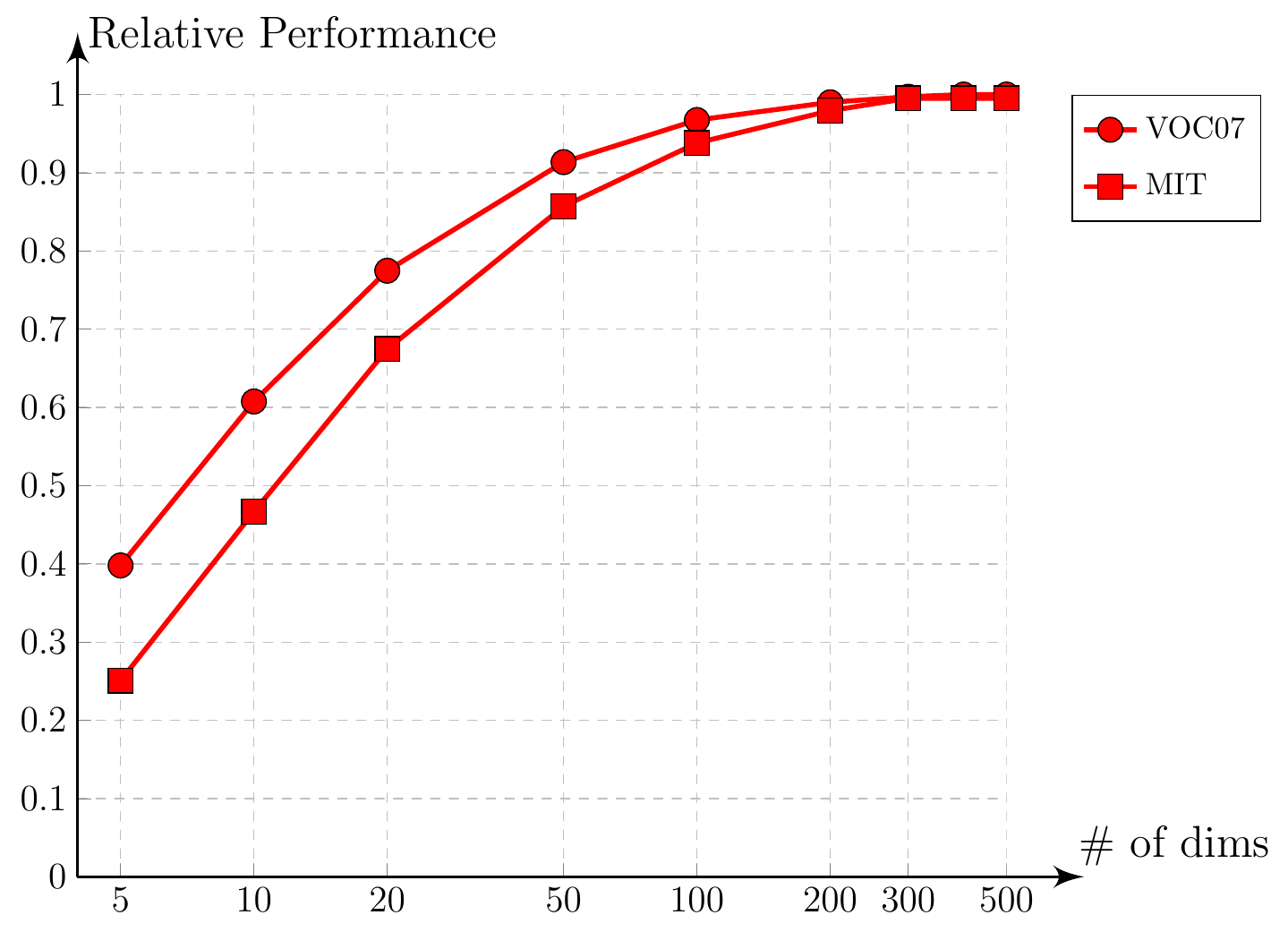}\includegraphics[width=0.5\linewidth]{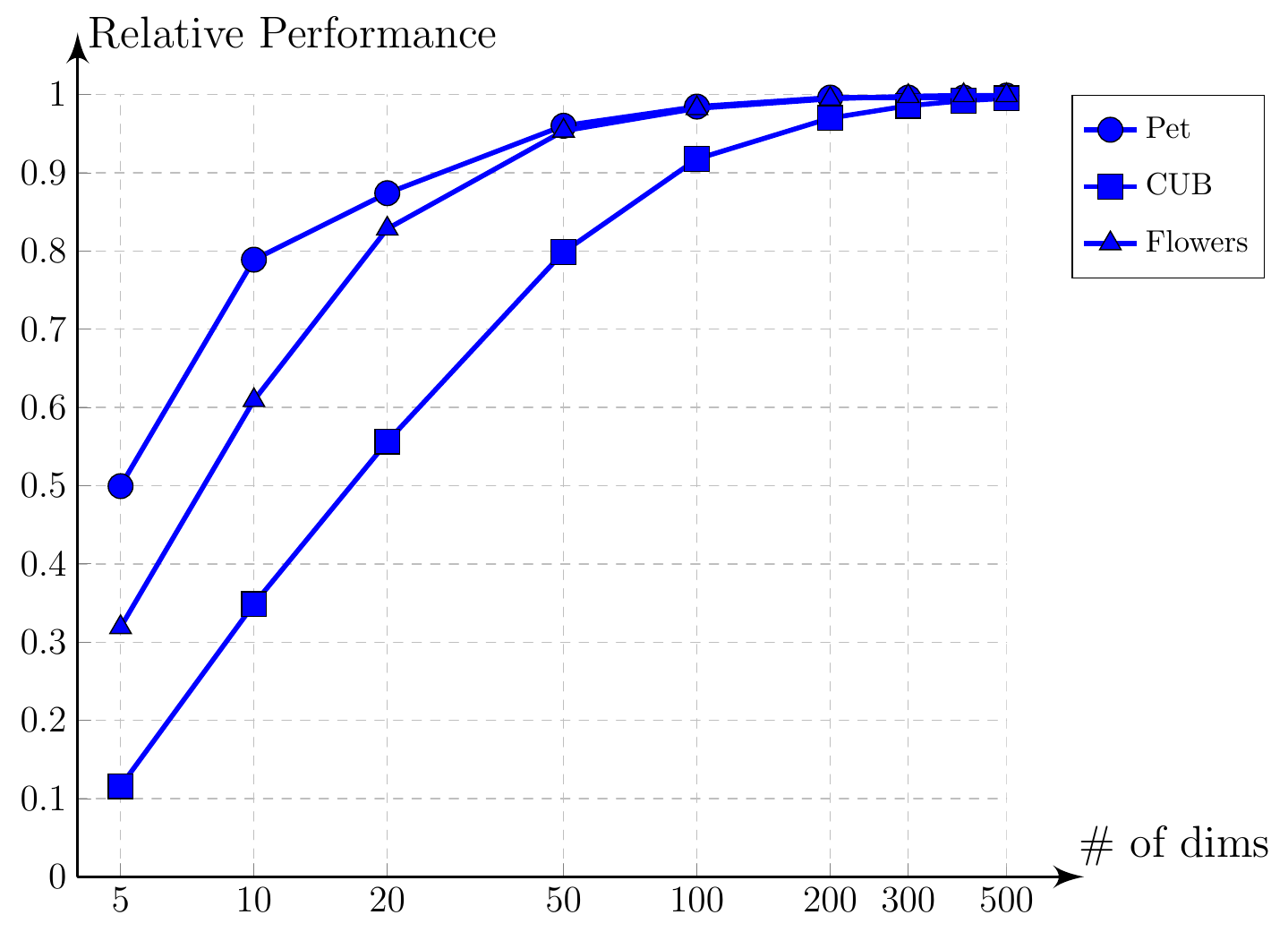}\\
   \includegraphics[width=0.5\linewidth]{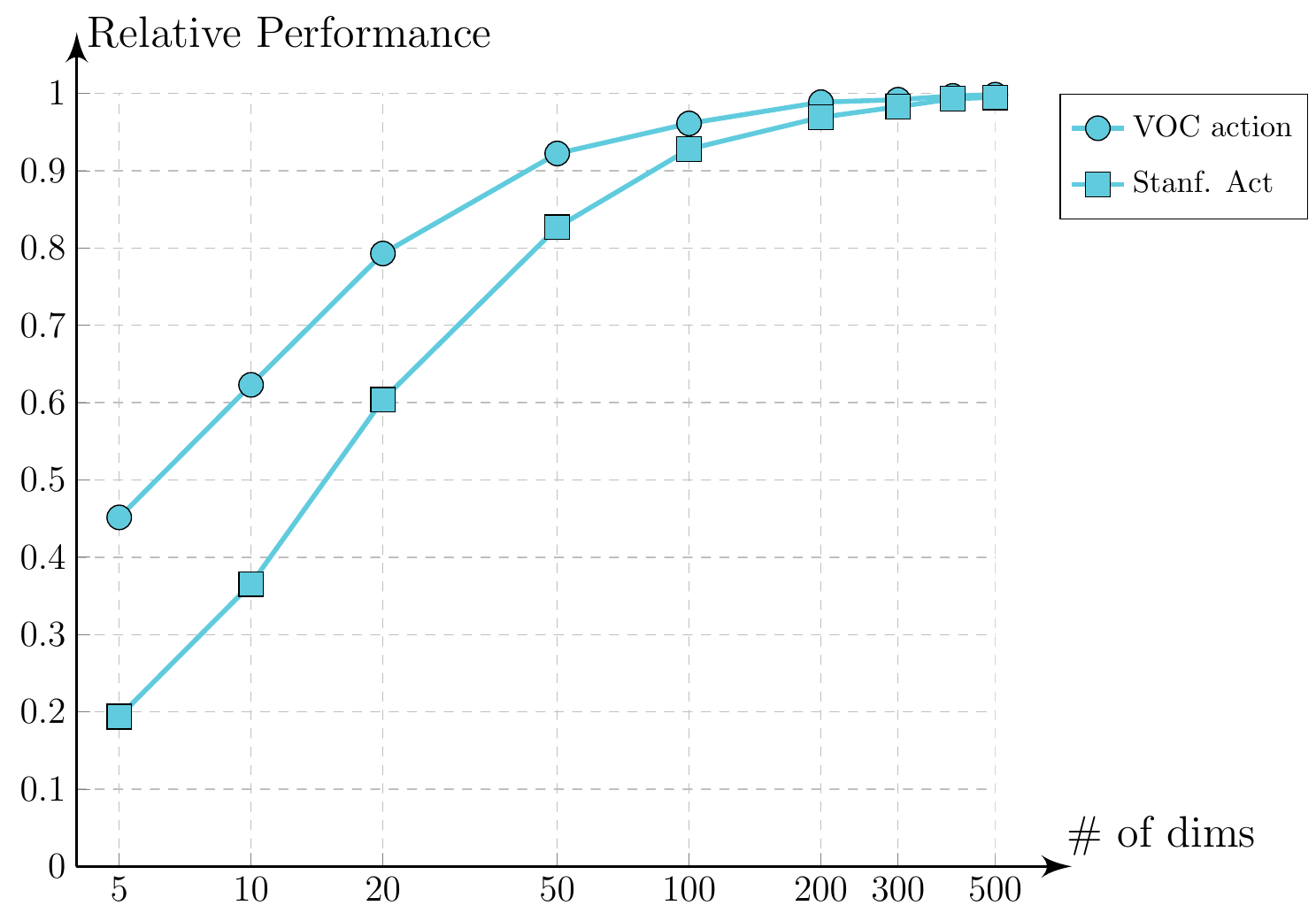}\includegraphics[width=0.5\linewidth]{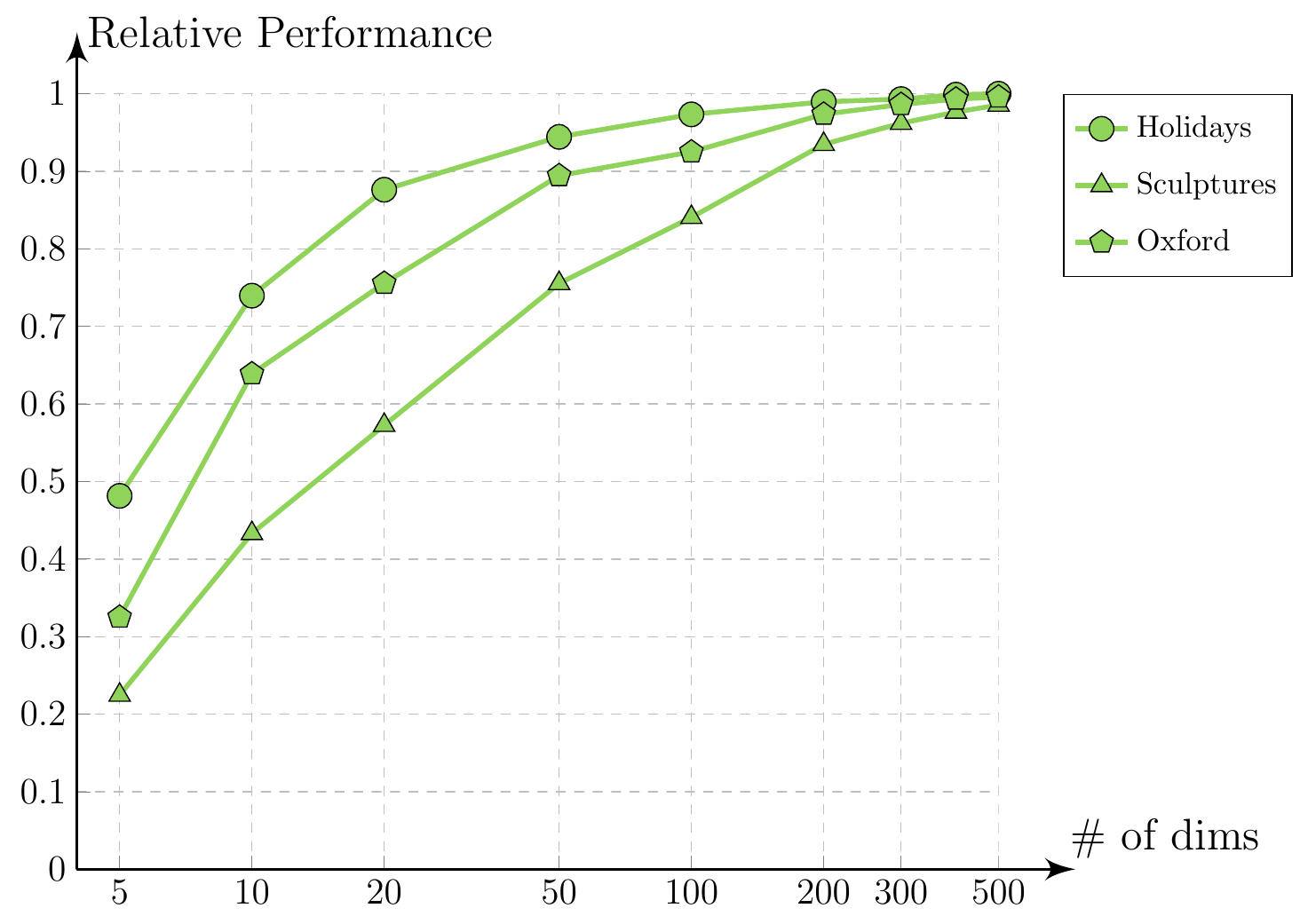}
\caption{\captionfontsize \textbf{Saturation:} As we move further from the original task, performance saturation happens with slightly more dimensions. For more clarity, the performance values of each task are divided by the maximum value for these plots. Horizontal axis is in log scale.}
\label{fig:pca2}
        \end{subfigure}%
\caption{\captionfontsize \textbf{Dimensionality Reduction:} We use Principal Component Analysis (PCA) to linearly transform the ConvNet representations obtained from first fully connected layer (4096 dimensional) into a lower dimensional space for various tasks. }
\end{figure*}

\subsubsection{Dimensionality Reduction}
\vspace{0.15cm}
We use principal component analysis (PCA) to reduce the dimensionality
of the transferred representation for each task. We observed that
dimensionality reduction helps all the instance retrieval tasks (most of the time insignificantly though). The main difference between the
retrieval task and other ones is that in retrieval we are interested
in the Euclidean distances between samples in the ConvNet representational space. In that respect, PCA can decrease the curse of dimensionality for \L2 distance. However, one could expect that dimensionality reduction would decrease the level of noise (and avoid potential over-fitting to irrelevant features for each specific task). But our experiments shows that this is not the case when using PCA for reducing dimensions. Figure \ref{fig:pca} shows the results for different tasks as we reduce the dimensionality of ConvNet representations. The results show that the relative performance boost gained by additional dimensions is correlated with the distance of the target task to the original task. We see that saturations appear earlier for the tasks closer to ImageNet. It is amazing to know that effective dimensionality of the ConvNet representations (with 4096 dims) used in these experiments is at most 500 for all visual recognition tasks from different domains. Another interesting observation is that many of the tasks work reasonably well with very low number of dimensions (5-50 dimensions). Remember that these features are obtained by a \textit{linear} transformation of the original ConvNet representation. This can indicate the capability of ConvNets in  linear factorization of the underlying generating factors of semantic visual concepts.\\

\begin{table}[]
\scriptsize
  \centering
      \begin{tabular}{l l l l}
        \topline
        \rowcolor{tableheadcolor}
        \textbf{Representation} &MIT&CUB&Flower\\ \midtopline \rowcolor{white} Medium FC7 &
        65.9 & 62.9 & 90.4\\ \midrule Medium FT & \textbf{66.3}
        & \textbf{66.4} & \textbf{91.4}\\
        \bottomline
      \end{tabular}
  \caption{\captionfontsize \textbf{Fine-tuning:} The first row shows the
    original ConvNet results. The second row shows the results when we
    fine-tune the ConvNet toward the target task and specialize the
    learnt representation. Fine-tuning is consistently effective. The
    proportional improvement is higher for the more distant tasks from
    ImageNet.}
\label{fig:fine_tuning}
\end{table}

\subsubsection{Fine-tuning}
\vspace{0.15cm}
\label{sec:fine_tuning}
Frequently the goal is to maximize the performance of a recognition
system for a specific task or a set of tasks. In this case intuitively
specializing the ConvNet to solve the task of interest would be the most
sensible path to take. Here we focus on the issue of fine-tuning the
ConvNet's representation with labelled data similar to those we expect to
see at test time.

\cite{Girshick13,Chatfield14} have shown that fine-tuning the network
on a target task helps the performance. Fine-tuning is done by
initializing a network with weights optimized for ILSVRC12. Then,
using the target task training set, the weights are updated. The
learning rate used for fine-tuning is typically set to be less than
the initial learning rate used to optimize the ConvNet for ILSVRC12. This
ensures that the features learnt from the larger dataset are not
forgotten. The step used to shrink the learning rate schedule is also
decreased to avoid over-fitting. We have conducted fine-tuning on the
tasks for which labels are mutually exclusive. The table
in Figure \ref{fig:fine_tuning} shows the results. The gains made by fine-tuning
increase as we move further away from the original image-level object
classification task. Fine-tuning on a relatively small target dataset
is a fast procedure. With careful selection of parameters it
is always at least marginally helpful.\\

\subsubsection{Increasing training data}
\vspace{0.15cm}
\label{sec:increase_data}
Zhu et al. \cite{Zhu12} suggest that increasing data is less effective
than increasing the complexity of models or richness of representation
and the former is prone to early performance saturation. Those
observations are made using HOG features to perform object
detection. Here, we want to investigate whether we are close to
saturation point with ConvNet representations.\\

\paragraph{Increasing data for target task.} To measure the effect of adding more data to learn the representation we consider the challenging task of PASCAL VOC 2007 object
detection. We follow the procedure of Girshick et al. \cite{Girshick13}
by fine-tuning the \alexnetA network using samples from the Oxford Pet
and Caltech-UCSD birds datasets. We show that although there exists a
large number of samples for those classes in ImageNet (more than
100,000 dogs) adding around $\sim$3000 dogs from the Oxford Pet
dataset helps the detection performance significantly. The same
improvement is observed for cat and bird, see the table in Figure
\ref{fig:fine_tuning}. This further adds to the evidence that
specializing a ConvNet representation by fine-tuning, even when the
original task contained the same labels, is helpful.\\

\paragraph{Increasing data for source task.} Furthermore, we investigate how important it is to increase training
data for the original ConvNet training. We train two networks, one
using SUN397 \cite{Xiao10} with 130K images and the other using the
Places dataset \cite{Zhou14} with 2.5M images. Then we test the
representations on the MIT Indoor Scenes dataset. The representation
trained from SUN397 (\textbf{62.6\%}) works significantly worse than
that of the Places dataset (\textbf{69.3\%}). The same trend is
observed for other datasets (refer to Table \ref{tab:additional_data_sun}). Since ConvNet representations can model very rich
representations by increasing its parameters, we believe we are still
far from saturation in its richness.

\begin{table}[t!]
\setlength{\tabcolsep}{3.9pt}
  \centering
    \centering
    \footnotesize
      \begin{tabular}{l r c l l l l l l}
        &\multicolumn{2}{c}{\textcolor{red}{\textbf{Classification}}} & \multicolumn{2}{c}{\textcolor{Plum}{\textbf{Attribute}}} & \multicolumn{2}{c}{\textcolor{blue}{\textbf{Fine-grained}}} &  \multicolumn{2}{c}{\textcolor{YellowGreen}{\textbf{Retrieval}}}\\
        \cmidrule[1pt](l{2em}r{2em}){2-3}
        \cmidrule[1pt](l{2em}r{2em}){4-5}
        \cmidrule[1pt](l{2em}r{2em}){6-7}
        \cmidrule[1pt](l{2em}r{2em}){8-9}
        \textbf  {Dataset} 			&VOC07			&MIT			&H3D			&UIUC			&Pet			&Flower			&Oxf.			&Scul.\\ 
        \midrule SUN	 			&57.8			&62.6			&45.0			&86.3			&45.0			&75.9			&64.5			&39.2\\ 
        \midrule Places				&\textbf{68.5}	&\textbf{69.3}	&\textbf{49.9}	&\textbf{88.8}	&\textbf{49.9}	&\textbf{82.4}	&\textbf{70.0}	&\textbf{44.2}\\
        \bottomline
      \end{tabular}
  \caption{\captionfontsize \textbf{Additional data (source task):} Results with the ConvNet representation optimized for different amount of training data. First row shows the results when the network is trained on scene recognition dataset of SUN397 \cite{Xiao14} dataset with ~100K images. The second row corresponds to the network trained on Places dataset \cite{Zhou14} with ~2.5M images annotated with similar categories. In all cases the ConvNet trained on Places dataset outperforms the one trained on SUN.}
\label{tab:additional_data_sun}
\end{table}
\begin{table*}[ht!]
\setlength{\tabcolsep}{1.1pt}
  \centering
    \centering
    \scriptsize
      \begin{tabular}{l l l l l l l l l l l l l l l l l l}
        &\multicolumn{3}{c}{\textcolor{red}{\textbf{Image Classification}}} & \multicolumn{3}{c}{\textcolor{Plum}{\textbf{Attribute Detection}}} & \multicolumn{3}{c}{\textcolor{blue}{\textbf{Fine-grained Recognition}}} & \multicolumn{3}{c}{\textcolor{SkyBlue}{\textbf{Compositional}}} & \multicolumn{5}{c}{\textcolor{YellowGreen}{\textbf{Instance Retrieval}}}\\
        \cmidrule[1pt](l{2em}r{2em}){2-4}
        \cmidrule[1pt](l{2em}r{2em}){5-7}
        \cmidrule[1pt](l{2em}r{2em}){8-10}
        \cmidrule[1pt](l{2em}r{2em}){11-13}
        \cmidrule[1pt](l{2em}r{2em}){14-18}
        \textbf  {}			 					&VOC07			&MIT			&SUN			&SunAtt			&UIUC			&H3D			&Pet			&CUB			&Flower			&VOCa.			&Act40			&Phrase			&Holid.			&UKB			&Oxf.			&Paris			&Scul.\\ 
		\midrule \multirow{2}{2pt}{non-ConvNet}&\cite{Song11}&\cite{Lin14}&\cite{Xiao14}&\cite{Patterson12}&\cite{Tsagkatakis10}&\cite{Zhang13}&\cite{Parkhi12}&\cite{Gavves13}&\cite{Koniusz13}&\cite{Oquab13}&\cite{Yao11_ICCV}&\cite{Sadeghi11}&\cite{Tolias13}&\cite{Zhao13}&\cite{Tolias13}&\cite{Tolias13}&\cite{Arandjelovic11}	\\											
        				&71.1		&68.5				&37.5			&87.5			&90.2			&69.1			&59.2			&62.7		&90.2			&69.6				&45.7		&41.5				&82.2			&89.4			&\textbf{81.7}	&78.2			&45.4\\ 
		\midrule Deep Standard					&71.8			&64.9			&49.6			&91.4			&90.6			&73.8			&78.5			&62.8			&90.5			&69.2			&58.9			&77.3			&86.2			&93.0		&73.0			&81.3			&53.7\\
        \midrule Deep Optimized\footnote[4]{}		 		&\textbf{80.7}	&\textbf{71.3}	&\textbf{56.0}	&\textbf{92.5}	&\textbf{91.5}	&\textbf{74.6}	&\textbf{88.1}	&\textbf{67.1}	&\textbf{91.3}	&\textbf{74.3}	&\textbf{66.4}	&\textbf{82.3}	&\textbf{90.0}	&\textbf{96.3}	&79.0		&\textbf{85.1}	&\textbf{67.9}\\
        \midrule Err. Reduction					&32\%			&18\%			&13\%			&13\%			&10\%			&4\%			&45\%			&12\%			&8\%			&17\%			&18\%			&22\%			&28\%			&47\%			&22\%			&20\%			&31\%\\
		\bottomline
        \midrule Source Task &ImgNet&Hybrid&Hybrid&Hybrid&ImgNet&ImgNet&ImgNet&ImgNet&ImgNet&ImgNet&ImgNet&ImgNet&Hybrid&ImgNet&ImgNet&ImgNet&ImgNet\\
        \midrule Network Width &Medium&Medium&Medium&Medium&Large&Medium&Medium&Medium&Medium&Medium&Medium&Medium&Medium&Medium&Medium&Medium&Medium\\
        \midrule Network Depth &16&8&8&8&8&16&16&16&16&16&16&16&8&8&16&16&16\\
        \midrule Rep. Layer &last&last&last&last&2nd last&2nd last&2nd last&3rd last&3rd last&3rd last&3rd last&3rd last&4th last&4th last&4th last&4th last&4th last\\
        \midrule PCA &\tickNo&\tickNo&\tickNo&\tickNo&\tickNo&\tickNo&\tickNo&\tickNo&\tickNo&\tickNo&\tickNo&\tickNo&\tickYes&\tickYes&\tickYes&\tickYes&\tickYes\\
        \midrule Pooling &\tickNo&\tickNo&\tickNo&\tickNo&\tickNo&\tickNo&\tickNo&\tickNo&\tickNo&\tickNo&\tickNo&\tickNo&$1\times1$&$1\times1$&$2\times2$&$2\times2$&$3\times3$\\
        \bottomline 
        \bottomline
      \end{tabular}
  \caption{\captionfontsize \textbf{Final Results:} Final results of the
    deep representation with optimized factors along with a linear SVM
    compared to the non-ConvNet state of the art. In the bottom half
    of the table the factors used for each task are noted. We achieve
    up to a \textbf{50\% reduction of error} by optimizing
    transferability factors. Relative error reductions refer to how
    much of the remaining error (from Deep Standard) is
    decreased. "Deep Standard" is the common choice of parameters - a
    Medium sized network of depth 8 trained on ImageNet with
    representation taken from layer 6 (FC6).}
\label{tab:final_results}
\end{table*}

\section{Optimized Results}
\label{sec:final_results}
In the previous section, we listed a set of factors which can affect the efficacy of the transformed representation from a generic ConvNet. We studied how best values of these factors are related to the distance of the target task to the ConvNet source task. Using the know-hows obtained from these studies, now we transfer the ConvNet representations using "Optimized" factors and compare the "Standard" ConvNet representation used in the field. The "Standard" ConvNet representation refers to a ConvNet of medium size and depth 8 (\alexnet) trained on 1.3M images of ImageNet, with the representation taken from first fully connected layer (FC6). As can be seen in Table \ref{tab:final_results} the remaining error of the "Standard" representation can be decreased by a factor of up to 50\% by optimizing its transferability factors.


\begin{table}[t!]
\centering
{\scriptsize
  \begin{tabular}{llll}
    \topline
    \rowcolor{tableheadcolor}
    \textbf{Representation} & bird & cat & dog\\
    \midtopline
    \rowcolor{white}
    ConvNet \cite{Girshick13} &38.5&51.4&46.0\\
    \midrule
    ConvNet-FT VOC \cite{Girshick13} &50.0&60.7&56.1\\
    \midrule
    ConvNet-FT VOC+CUB+Pet&\textbf{51.3}&\textbf{63.0}&\textbf{57.2}\\
    \bottomline
  \end{tabular}
}
\caption{\captionfontsize \textbf{Additional data (fine-tuning):} The table presents
  the mAP accuracy of a sliding window detector based on different
  ConvNet representations for 3 object classes from VOC 2007.
  ImageNet contains more than 100,000 dog images and Pascal VOC has
  510 dog instances. For the representation in the second row, image
  patches extracted from the VOC training set are used to fine-tune
  the ConvNet representation\cite{Girshick13}. It results in a big
  jump in performance. But including cat, dog and bird images from the
  Oxford Pet and Caltech bird datasets boosts the performance even
  further.}
\label{tab:additional_data}
\end{table}

\section{Implementation details}
\label{sec:implementation}
The \caffeA software \cite{Jia14} is used to train our
ConvNets. Liblinear is used to train the SVMs we use for
classification tasks. Retrieval results are based on the \L2 distance
of whitened ConvNet representations. All parameters were selected
using 4-fold cross-validation. \\

Learning choices are the same as \cite{Razavian14}. In particular, the pipeline for classification tasks is as follows: we first construct the feature vector by getting the average ConvNet feature vector of 12 jittered samples of the original image. The jitters come from crops of 4 corners of the original image, its center and the whole image resized to the size needed by the network (227x227) and their mirrors. We then \L2 normalize the ConvNet feature vector, raise the absolute value of each feature dimension to the power of 0.5 and keep its sign. We use linear SVM trained using one-versus-all approach for multilabel tasks (e.g. PASCAL VOC image classification) and linear SVM trained using one-versus-one approach and voting for single label tasks (e.g. MIT Indoor Scene). \\

The pipeline for the retrieval tasks are as follows: Following \cite{JegouECCV12} The feature vectors are first \L2 normalized, then the dimensionality is reduced using PCA to smaller whitened dimension and the resulting feature is renormalized to the unit length. Since buildings (Oxford and Paris) and scupltures datasets include partial images or the object can appear in small part of the whole image (zoomed in or out images of the object of interest) we use spatial search to match windows from each pair of images. We have 1 sub-patch of size 100\% of the whole image, 4 sub-patches of each covering 4/9 size of the image. 9 sub-patches of each covering 4/16 and 16 sub-patches of each covering 4/25 of the image (in total 30 sub-paches). The minimum distance of all sub-patches is considered as the distance of the two images.\\

\section{Closing Discussion}
\label{sec:discussion}

ConvNet representations trained on ImageNet are becoming the standard
image representation. In this paper we presented a systematic study,
lacking until now, of how to effectively transfer such representations
to new tasks. The most important elements of our study are: We
identify and define several factors whose settings affect
transferability. Our experiments investigate how relevant each of
these factors is to transferability for many visual recognition
tasks. We define a categorical grouping of these tasks and order them
according to their distance from image classification.

Our systematic experiments have allowed us to achieve the
following. First, by optimizing the identified factors we improve the
state-of-the-art performance on a very diverse set of standard
computer vision databases, see table \ref{tab:final_results}. Second,
we observe and present empirical evidence that the effectiveness of a
factor is highly correlated with the distance of the target task from
the source task of the trained ConvNet. Finally, we empirically verify
that our categorical grouping and ordering of visual recognition tasks
is meaningful as the optimal setting of the factors remain constant
within each group and vary in a consistent manner across our
ordering. Of course, there are exceptions to the general trend. In
these few cases we provide simple explanations.

We think the insights generated by our paper can be used to learn more
generic features (our ultimate goal). We believe a generic visual
representation must encode different levels of visual information
(global, local and visual relations) and invariances. Although these
levels of information and invariances are interconnected, a task can
be analyzed based on which level of information it requires. And this
allows us to explain the distance of visual recognition tasks from
that of ImageNet and then crucially to identify orthogonal
training tasks that should be combined when training a generic
representation. Because when we optimize a representation for only one
type of invariance and/or visual information we cannot expect it to
optimally encode the others.

During ConvNet training it is the loss function, besides the semantic
labels, that controls the learnt representation. For example for image
classification we want different semantic classes to occupy
non-overlapping volumes of the representation space. The cross-entropy
loss function promotes this behaviour. While if we want to learn a
representation to measure visual similarity we must use a different
loss function as we also need the representation of images with the
same label to occupy a small volume.

Therefore, in future work, we plan to investigate how to best apply
multi-task learning with ConvNets to learn generic representations.
We will focus on how to choose the training tasks and loss functions
that will force the ConvNet representation to learn many different
levels of visual information incorporating different levels of
invariances.

\section{Acknowledgement}
We gratefully acknowledge the support of NVIDIA Corporation with the donation of the Tesla K40 GPUs to this research.
 \footnotetext[4]{\textbf{ Note:} "Deep Optimized" results in this table are not always the optimal choices of factors studied in the paper. For instance one would expect a very deep network trained using hybrid model would improve results on MIT and SUN, or a deep and large network would perform better on VOC image classification. Another example is that we could do fine-tuning with the optimal choices of parameters for nearly all tasks. Obviously, it was highly computationally expensive to produce all the existing results. We will update the next versions of the paper with further optimized choices of parameters.}
{
\bibliographystyle{ieee}
\bibliography{local}
}
\end{document}